\documentclass[journal]{IEEEtran}
\usepackage[T1]{fontenc}
\usepackage{graphicx}
\usepackage{times}
\usepackage{helvet}
\usepackage{courier}
\usepackage{amsmath}
\usepackage{algorithm}
\usepackage{algorithmic}
\usepackage{csquotes} 
\usepackage{color}
\usepackage{paralist}
\usepackage{amssymb}
\usepackage{indentfirst}
\usepackage{subfigure}
\usepackage{float}
\usepackage{multirow}
\usepackage{cite}
\usepackage{mathrsfs}

\usepackage{mathrsfs} 
\usepackage{amsfonts}
\usepackage{todonotes}
\usepackage{pgfplots} 
\usepackage{epstopdf}
\usepackage{epsfig}
\pgfplotsset{compat=newest}
\usepackage{color}
\definecolor{forestgreen}{RGB}{0,139,69}

\usepackage{xcolor}
\definecolor{citecolor}{HTML}{0071bc}
\usepackage[colorlinks, linkcolor=red,  anchorcolor=blue, citecolor=citecolor]{hyperref} 

\usepackage{xcolor}
\definecolor{SeaGreen4}{RGB}{0,205,102} 
\definecolor{SlateBlue}{RGB}{106,90,205} 
\definecolor{DarkRed}{RGB}{178,34,34} 
	
\usepackage[switch]{lineno}

\usepackage{textcomp,booktabs}
\usepackage{amssymb}
\usepackage{pifont}
\newcommand{\cmark}{\ding{51}}%

\usepackage{makecell}

\usepackage{colortbl}
\definecolor{mygray}{gray}{.9}
\definecolor{mypink}{rgb}{.99,.91,.95}
\definecolor{mycyan}{cmyk}{.3,0,0,0}

\begin{document}

\title{ Dynamic Pondering Sparsity-aware Mixture-of-Experts Transformer for Event Stream based Visual Object Tracking }

\author{Shiao Wang, Xiao Wang*, \emph{Member, IEEE}, Duoqing Yang, Wenhao Zhang, Bo Jiang*, Lin Zhu, \\ 
    Yonghong Tian, \emph{Fellow, IEEE}, Bin Luo, \emph{Senior Member, IEEE} 

\thanks{$\bullet$ Shiao Wang, Duoqing Yang, Wenhao Zhang, Xiao Wang, Bo Jiang, Bin Luo are with the School of Computer Science and Technology, Anhui University, Hefei 230601, China. (email: \{e24101001, e125221163\}@stu.ahu.edu.cn, \{xiaowang, jiangbo, luobin\}@ahu.edu.cn)} 

\thanks{$\bullet$ Lin Zhu is with Beijing Institute of Technology, Beijing, China. (email: linzhu@pku.edu.cn)}

\thanks{$\bullet$ Yonghong Tian is with Peng Cheng Laboratory, Shenzhen, China; National Key Laboratory for Multimedia Information Processing, School of Computer Science, Peking University, China; School of Electronic and Computer Engineering, Shenzhen Graduate School, Peking University, China. (email: yhtian@pku.edu.cn) }

\thanks{* Corresponding Author: Xiao Wang, Bo Jiang} 
}

\markboth{ IEEE Transactions on ***, 2026 } 
{Shell \MakeLowercase{\textit{et al.}}: Bare Demo of IEEEtran.cls for IEEE Journals}

\maketitle

\begin{abstract}
Despite significant progress, RGB-based trackers remain vulnerable to challenging imaging conditions, such as low illumination and fast motion. Event cameras offer a promising alternative by asynchronously capturing pixel-wise brightness changes, providing high dynamic range and high temporal resolution. 
However, existing event-based trackers often neglect the intrinsic spatial sparsity and temporal density of event data, while relying on a single fixed temporal-window sampling strategy that is suboptimal under varying motion dynamics. 
In this paper, we propose an event sparsity-aware tracking framework that explicitly models event-density variations across multiple temporal scales. Specifically, the proposed framework progressively injects sparse, medium-density, and dense event search regions into a three-stage Vision Transformer backbone, enabling hierarchical multi-density feature learning. Furthermore, we introduce a sparsity-aware Mixture-of-Experts module to encourage expert specialization under different sparsity patterns, and design a dynamic pondering strategy to adaptively adjust the inference depth according to tracking difficulty. 
Extensive experiments on FE240hz, COESOT, and EventVOT demonstrate that the proposed approach achieves a favorable trade-off between tracking accuracy and computational efficiency.
The source code will be released on \url{https://github.com/Event-AHU/OpenEvTracking}. 
\end{abstract}


\IEEEpeerreviewmaketitle

\section{Introduction}

\IEEEPARstart{V}{isual} Object Tracking (VOT) aims to continuously estimate the location of a target in subsequent frames, given its initial bounding box in the first frame. Most existing trackers~\cite{chen2021transt, wang2021TNL2K, yao2025unctrack, chen2025hyperspectral, wang2025ssf} are built upon traditional RGB imaging technology and have achieved remarkable success across a wide range of applications, including autonomous driving, surveillance, and drone photography. However, RGB cameras are inherently sensitive to challenging imaging conditions, such as overexposure, low illumination, motion blur, and fast motion, which often result in severe performance degradation or even tracking failure. These limitations substantially hinder the deployment of robust RGB-only trackers in complex real-world scenarios.

Bio-inspired event cameras offer an alternative sensing paradigm by asynchronously capturing pixel-wise brightness changes. Specifically, each pixel independently monitors the logarithmic intensity of incoming light and triggers an event once the intensity variation exceeds a predefined threshold. Each event is represented as $e=(x,y,t,p)$, where $(x, y)$ denotes the spatial location, $t$ is the timestamp, and $p \in \{+1, -1\}$ indicates the polarity of brightness increase or decrease. Owing to their event-driven acquisition mechanism, event cameras offer high temporal resolution, high dynamic range, low latency, and inherent privacy preservation, making them a promising complement or alternative to conventional RGB cameras, particularly in challenging imaging conditions~\cite{gallego2020event}.

\begin{figure}
\center
\includegraphics[width=0.95\linewidth]{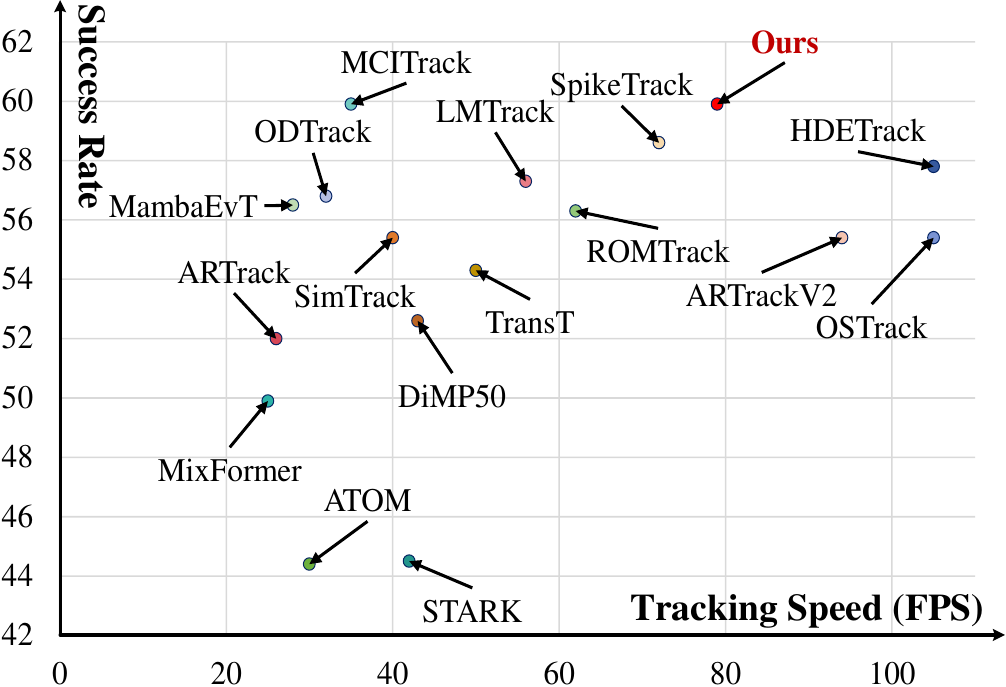}
\caption{Comparison with other trackers in terms of accuracy (Success Rate) and efficiency (FPS).} 
\label{SR&FPS}
\end{figure}

\begin{figure*}
\center
\includegraphics[width=\linewidth]{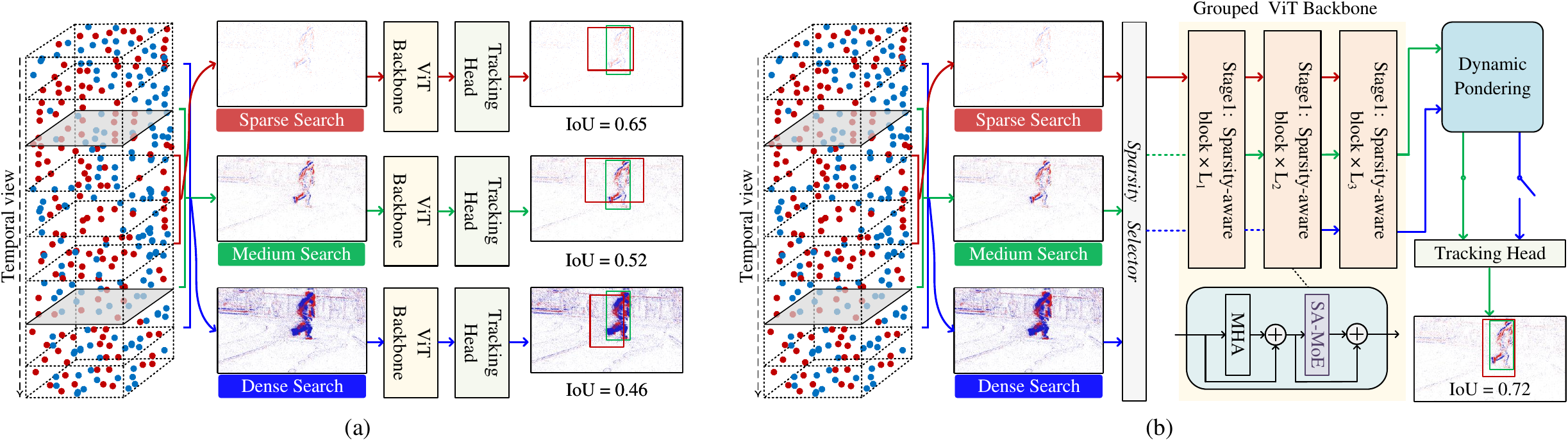}
\caption{(a) Different temporal windows produce event representations with varying densities, affecting the tracking results. Sparse backgrounds are often highly distinguishable from the target, and such relatively simple scenarios should be allocated fewer computational resources to improve tracking efficiency.
(b) Overview of the proposed Dynamic Pondering Sparsity-aware Mixture-of-Experts Transformer for efficient event-based tracking.} 
\label{firstIMG}
\end{figure*}

Motivated by the distinctive advantages of event cameras, recent studies~\cite{mambaevt2025, zhang2021object, wang2024event, zhu2022learning, zhang2026efficient} have investigated event-based tracking methods, aiming to exploit the high temporal resolution and high dynamic range of event streams for robust visual object tracking in challenging scenarios.
For example, Wang et al.~\cite{mambaevt2025} introduce MambaEVT, an event stream-based visual object tracking framework that leverages state space models to reduce the computational complexity of the model. Zhang et al.~\cite{zhang2021object} proposed STNet, a spiking-based tracking framework that exploits the asynchronous nature of event streams to capture rich spatiotemporal information, thereby improving tracking performance under challenging motion conditions. Wang et al.~\cite{wang2024event} further introduce cross-modal knowledge distillation, transferring complementary knowledge from multimodal data to the event domain to enhance tracking efficiency. 


Despite these advances, existing methods still struggle to achieve a favorable trade-off between tracking accuracy and computational efficiency, as shown in Fig.~\ref{SR&FPS}. This limitation mainly arises from three aspects. \textit{1). Insufficient temporal-scale utilization}: existing event-based trackers commonly rely on a fixed temporal window to construct event representations, leaving the influence of different temporal scales insufficiently utilized. As shown in Fig.~\ref{firstIMG} (a), short and long temporal windows preserve contour details and motion cues differently, making a single temporal scale insufficient to fully characterize event streams under varying motion dynamics. \textit{2). Underexplored spatial sparsity}: the inherent spatial sparsity of event data remains underexploited, resulting in limited sparsity-specific feature modeling. \textit{3). Redundant backbone inference}: existing trackers typically allocate computation uniformly across the backbone network regardless of scenario difficulty, leading to redundant inference for simple scenarios.

To address these limitations, we propose PSMTrack, a dynamic pondering sparsity-aware Mixture-of-Experts Transformer for efficient event-based visual object tracking. PSMTrack explicitly models multi-scale event-density variations, enhances sparsity-aware representations through expert specialization, and adaptively adjusts inference depth to achieve a favorable trade-off between tracking accuracy and computational efficiency.
As illustrated in Fig.~\ref{firstIMG} (b), sparse, medium-density, and dense search regions are jointly fed into a vision Transformer (ViT)~\cite{dosovitskiy2020image} backbone for feature learning. Specifically, the proposed framework organizes the ViT backbone into three stages and progressively injects search regions with different sparsity levels into deeper stages, enabling hierarchical multi-density feature modeling while reducing the computational overhead introduced by multi-sparsity inputs.
Furthermore, we introduce a sparsity-aware Mixture-of-Experts (MoE) module into the feed-forward networks of Transformer blocks. This design encourages different experts to specialize in modeling feature representations with varying sparsity levels, thereby enhancing representation capacity and feature discriminability. Finally, we propose a dynamic pondering strategy that adaptively determines whether to terminate inference at intermediate layers according to the difficulty of different tracking scenarios. This mechanism dynamically adjusts the inference depth, improving tracking efficiency while maintaining robust tracking performance.


To sum up, the contributions of this work can be summarized as follows:

\textit{1).} We propose a unified multi-sparsity modeling framework that captures event-density variations across temporal scales through hierarchical feature learning, thereby effectively exploiting the inherent spatial sparsity and temporal density of event data.

\textit{2).} We introduce a sparsity-adaptive tracking paradigm that integrates a sparsity-aware Mixture-of-Experts module with a dynamic pondering strategy, enabling efficient and robust tracking under diverse motion dynamics.

\textit{3).} Extensive experiments on three public benchmarks, i.e., FE240hz, COESOT, and EventVOT, demonstrate the effectiveness and efficiency of the proposed approach.

\section{Related Works} 

In this section, we review the most related research topics to our paper, including Event-based tracking, Efficient Transformer Tracking, and Mixture of Experts Network. More related works can be found in the following surveys~\cite{gallego2020event, marvasti2021trackSurvey} and paper list \footnote{\url{github.com/wangxiao5791509/Single_Object_Tracking_Paper_List}}.

\subsection{Event-based Tracking}
By asynchronously capturing pixel-level brightness changes, event cameras have attracted considerable attention due to their unique advantages, including high temporal resolution, low latency, and high dynamic range~\cite{gallego2020event}. These properties make them particularly well-suited for visual tracking in challenging scenarios, such as motion blur and low-light conditions, where conventional RGB cameras often struggle.

The early event-based tracker ESVM~\cite{8368143} was proposed by Huang et al. for high-speed moving object tracking.
Chen et al.~\cite{chen2019asynchronous} introduced ATSLTD, an adaptive event-to-frame conversion algorithm designed for asynchronous tracking.
EKLT~\cite{gehrig2020eklt} combined frame and event streams for high-temporal-resolution feature tracking.
Recently, deep learning architectures have shown their advantages in tracking tasks.
STNet~\cite{zhang2022spiking} integrated a Transformer and an SNN for spatiotemporal modeling. 
Zhu et al.~\cite{zhu2022learning} improved tracking accuracy and speed through key event sampling and graph network embedding.
Wang et al.~\cite{wang2024event} proposed a novel hierarchical cross-modality knowledge distillation approach.
Zhang et al.~\cite{zhang2026efficient} propose an efficient Vision Transformer architecture that reduces computational cost while maintaining comparable performance for event-based tracking.
Wang et al.~\cite{wang2025towards} propose a novel paradigm for event stream-based tracking that integrates a slow, high-accuracy tracker with a fast, low-latency tracker to enhance both efficiency and effectiveness.
In this work, we advance event-based tracking with a sparsity-aware Mixture-of-Experts Transformer that effectively exploits the spatial sparsity and temporal density of event data. We further design a dynamic pondering strategy to adaptively adjust the inference depth according to tracking difficulty, thereby improving tracking efficiency.

\subsection{Efficient Transformer Tracking}
Transformer-based efficient tracking algorithms have recently received increasing attention, as they aim to balance the powerful modeling capability of Transformers with the practical constraints of model complexity and inference speed. Early Transformer-based trackers, such as TransT~\cite{chen2021transt} and STARK~\cite{yan2021stark}, demonstrated the effectiveness of attention mechanisms in modeling search-template interactions. However, the quadratic complexity of self-attention with respect to the token length limits its efficiency, especially in high-resolution or long-term tracking scenarios.

To mitigate this issue, numerous RGB-based trackers~\cite{ye2022ostrack, zhu2025two, hong2025general, wang2024event, xu2025less, xue2025similarity, xu2025less} have focused on lightweight architectures and token pruning strategies. OSTrack~\cite{ye2022ostrack} introduces a candidate elimination strategy to remove irrelevant background regions, thereby improving efficiency. LightTrack~\cite{yan2021lighttrack} leverages neural architecture search to design compact Transformer-based trackers with reduced computational cost. MixFormerV2~\cite{cui2023mixformerv2} improves tracking efficiency through distillation-based backbone reduction. 
Zhu et al.~\cite{zhu2025exploring} achieve efficient object tracking via intermediate terminating branches.
Hong et al.~\cite{hong2025general} proposed CompressTracker, whose replacement training strategy improves the student model’s ability to mimic the teacher model while simplifying the training process. 

For event-based tracking, HDETrack~\cite{wang2024event} achieves high-speed visual tracking using only event streams during inference via a hierarchical knowledge distillation framework. 
Wang et al.~\cite{wang2025towards} improve both efficiency and performance by combining a slow, high-accuracy tracker with a fast, low-latency tracker. 
Zhang et al.~\cite{zhang2026efficient} further reduce computational cost through two adaptive token sparsification strategies tailored to event data and tracking-specific characteristics.
Inspired by A-ViT~\cite{yin2022vit}, we propose a dynamic pondering strategy that adaptively adjusts the inference depth of ViT layers according to tracking difficulty. Unlike conventional dynamic inference methods~\cite{yin2022vit, zhu2025exploring}, our strategy is tailored to event-based tracking by explicitly considering the spatial sparsity and temporal density of event streams, thereby achieving a favorable trade-off between tracking performance and computational efficiency.

\subsection{Mixture-of-Experts Network}
Mixture-of-Experts (MoE) Network has emerged as an effective paradigm for improving model capacity and computational efficiency by dynamically routing tokens to a subset of expert networks. The sparsely gated MoE layer proposed in~\cite{shazeer2017outrageously} demonstrates that large model capacity can be achieved with limited computation by activating only a few experts for each input. Subsequent studies, such as Switch Transformer~\cite{fedus2022switch}, further simplify the routing mechanism and improve training stability, making MoE more practical for large-scale applications.

Recent studies have extended MoE to vision tasks, particularly in combination with Vision Transformers. V-MoE~\cite{riquelme2021scaling} incorporates sparse expert layers into ViT and demonstrates improved scalability for image classification. 
Tutel~\cite{hwang2023tutel} focuses on system-level optimization, significantly improving the efficiency of MoE training and inference on modern hardware.
Moreover, several works investigate advanced routing strategies and expert utilization. 
Lewis et al.~\cite{lewis2021base} introduce a balanced assignment mechanism to alleviate expert load imbalance. 
Zoph et al.~\cite{zoph2022stmoe} propose stability-enhanced routing and training strategies for sparse expert models. 
Roller et al.~\cite{roller2021hash} adopt hash-based routing to reduce routing overhead.
Lu et al.~\cite{lu2025dynamic} integrate MoE to achieve real-time open-vocabulary object detection.
Zhao et al.~\cite{zhao2025equipping} apply MoE to out-of-distribution detection in visual foundation models.
In this work, we propose a sparsity-aware MoE module for event-based visual tracking, in which multiple sub-experts are designed to specialize in event representations with different sparsity levels. This design enables flexible expert specialization and improves adaptation to the inherent sparsity characteristics of event data.

\section{Our Proposed Approach} 

\subsection{Overview}
To fully exploit the spatial sparsity and temporal density of event data while improving tracking efficiency, we propose PSMTrack, as illustrated in Fig.~\ref{framework}. Specifically, based on short-, medium-, and long-term temporal windows, we construct three event search regions with different sparsity levels instead of relying on a single fixed temporal window. This design enables the model to capture sharp object contours from short-term observations and rich motion dynamics from long-term information.
Building upon this design, we divide the Vision Transformer backbone into three stages to progressively incorporate event features with varying sparsity levels. Meanwhile, we introduce a sparsity-aware MoE module into the feed-forward network of the Transformer block, encouraging different experts to specialize in features with different sparsity levels, thereby enhancing feature discriminability across diverse event densities.
To further improve efficiency, as shown in Fig.~\ref{Sub_Fig}, we propose a dynamic pondering strategy that adaptively adjusts the inference depth according to tracking difficulty, achieving a favorable trade-off between tracking accuracy and computational cost. The overall architecture will be detailed in the following sections, including the sparsity-aware Transformer block and the dynamic pondering strategy.

\subsection{Input Representation}

\begin{figure}
\includegraphics[width=0.98\columnwidth]{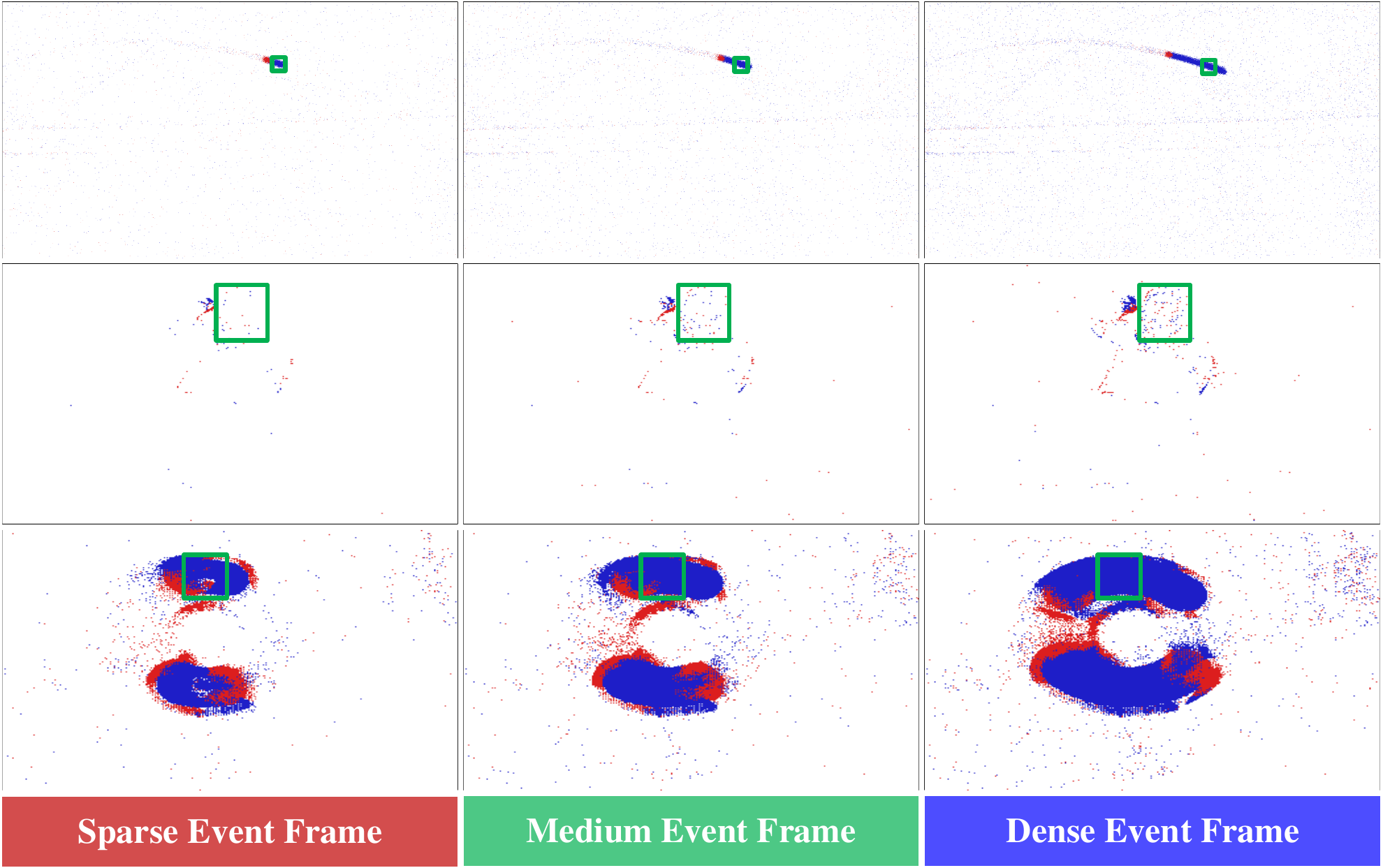}
\caption{Visualization of event representations with different event densities.}  
\label{Sparsity_Visualization}
\end{figure}

Given an event stream $\mathcal{E}_P=\{e_i\}_{i=1}^{M}$ within a time interval $T$, consisting of $M$ asynchronously generated events, each event $e_i$ is represented as a quadruple $(x_i,y_i,t_i,p_i)$. Here, $(x_i,y_i)$ denotes the spatial coordinates, $t_i\in[0,T]$ is the timestamp, and $p_i\in\{-1,1\}$ indicates the polarity. The event stream is partitioned into $N=T/\Delta t$ temporal segments, where $\Delta t$ denotes the duration of each segment. As observed in~\cite{zhu2025separation} and Fig.~\ref{Sparsity_Visualization}, event frames constructed from short temporal windows preserve sharper edge details, whereas those generated from long temporal windows encode richer motion cues.
For the $k$-th segment centered at time $t_k$, we construct three temporal windows at different scales, denoted as $\mathcal{W}_s^{k}$, $\mathcal{W}_m^{k}$, and $\mathcal{W}_d^{k}$:
\begin{equation}
\begin{aligned}
\mathcal{W}_s^{k} &= \left[t_k-\tfrac{1}{4}\Delta t,\, t_k+\tfrac{1}{4}\Delta t\right], \\
\mathcal{W}_m^{k} &= \left[t_k-\tfrac{1}{2}\Delta t,\, t_k+\tfrac{1}{2}\Delta t\right], \\
\mathcal{W}_d^{k} &= \left[t_k-\tfrac{3}{4}\Delta t,\, t_k+\tfrac{3}{4}\Delta t\right],
\end{aligned}
\end{equation}
where the subscripts $s$, $m$, and $d$ denote sparse, medium-density, and dense temporal windows, respectively. These windows are used to generate sparse, medium-density, and dense spatiotemporal event representations. Accordingly, we obtain the stacked event representation $\mathcal{E}\in\mathbb{R}^{3N\times3\times H\times W}$, where the second dimension denotes the channel number, and $H$ and $W$ represent the height and width of each event frame, respectively.

\subsection{Network Architecture}

\noindent $\bullet$ \textbf{Three-Stage Vision Transformer Backbone.}
Given event frames with multiple sparsity levels, we adopt the medium-density representation of the first frame as the template to provide a balanced reference. For subsequent frames, event representations at all three density levels are used as search regions.
All inputs are cropped and resized to fixed spatial resolutions, yielding the template frame $Z_0 \in \mathbb{R}^{3 \times H_z \times W_z}$ and the search frames $\{X_s^i, X_m^i, X_d^i\}$, where $X_s^i, X_m^i, X_d^i \in \mathbb{R}^{3 \times H_x \times W_x}$ correspond to sparse, medium-density, and dense search inputs, respectively. The cropped event frames are then partitioned into non-overlapping patches of size $P \times P$. Each patch is projected into a $D$-dimensional embedding space through a linear projection layer. After adding positional encodings, we obtain the template tokens $Z \in \mathbb{R}^{N_z \times D}$ and the search tokens $\{X_s, X_m, X_d\}$, where $X_s, X_m, X_d \in \mathbb{R}^{N_x \times D}$. Here, $N_z=H_zW_z/P^2$ and $N_x=H_xW_x/P^2$ denote the numbers of template and search tokens, respectively.

\begin{figure*}
\center
\includegraphics[width=0.95\linewidth]{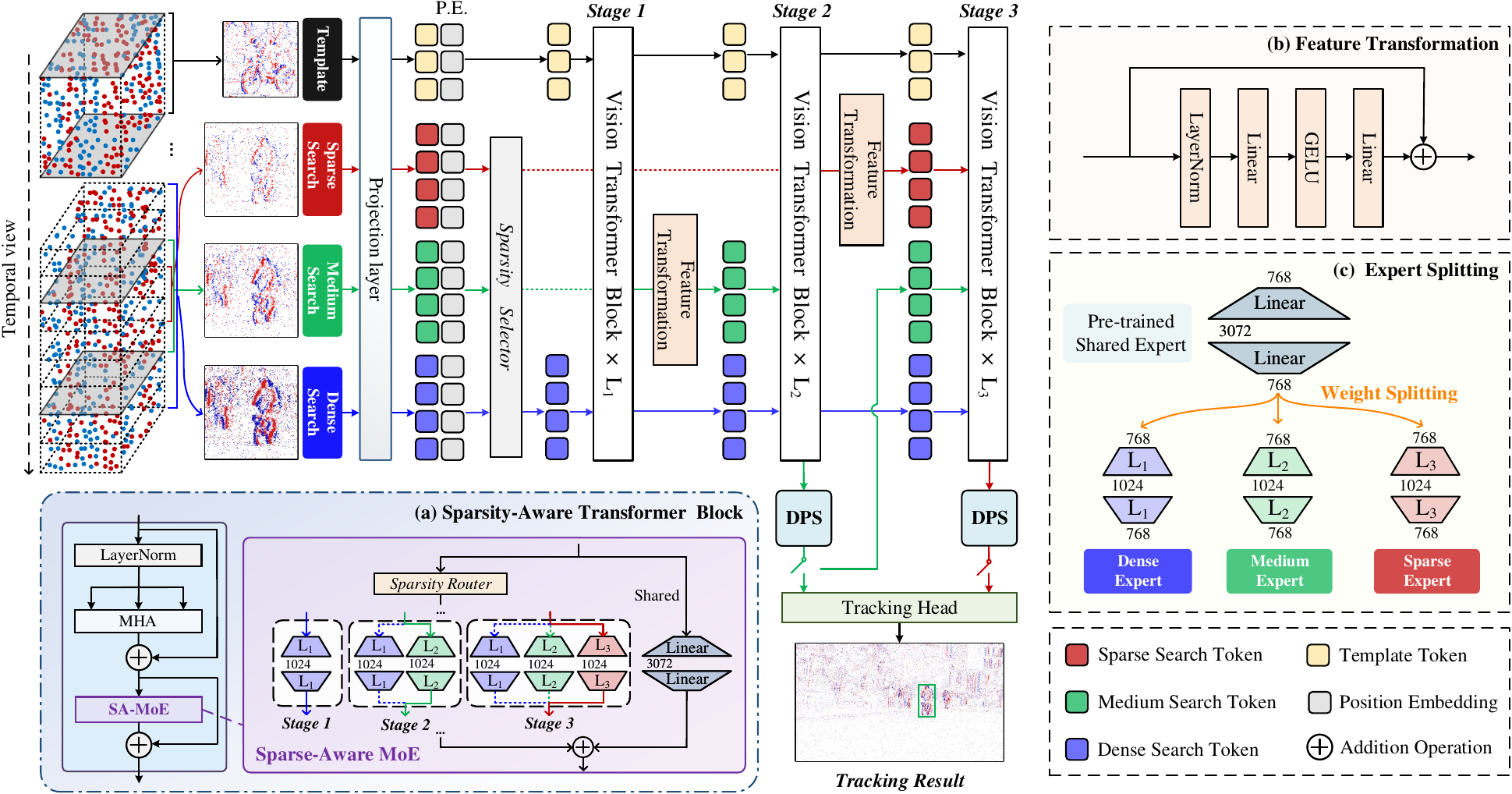}
\caption{The overall framework of the proposed Dynamic \textbf{P}ondering \textbf{S}parsity-aware \textbf{M}ixture-of-Experts Transformer for event-based tracking, termed PSMTrack. According to different temporal window lengths, sparse, medium-density, and dense event representations are jointly fed into a hierarchical backbone network for progressive feature learning. Specifically, we introduce a sparsity-aware Mixture-of-Experts (MoE) module into the first block of each stage to replace the standard feed-forward network, enabling specialized modeling of feature representations with different sparsity levels. In addition, we propose a dynamic pondering strategy to adaptively determine whether to terminate inference early, thereby improving overall tracking efficiency.}
\label{framework}
\end{figure*}

Although multi-sparsity search representations provide complementary event cues, naively concatenating all search tokens with the template tokens and feeding them into the entire Transformer backbone would substantially increase the computational cost due to the quadratic complexity of self-attention. To address this issue, we divide the Transformer backbone into three stages, containing $\{6,4,2\}$ layers in our final implementation, and design a multi-stage feature learning strategy.
Specifically, \textbf{in the first stage}, a search region with a specific sparsity level is selected as the initial input. Taking the dense search region as an example, we concatenate it with the template tokens to obtain $F_d \in \mathbb{R}^{(N_z+N_x)\times D}$, which is then fed into the first-stage Vision Transformer blocks to model template-search interactions.
\textbf{In the second stage}, another search region with a different sparsity level, e.g., the medium-density search region, is incorporated into the token sequence, forming $F_{d,m} \in \mathbb{R}^{(N_z+2N_x)\times D}$. Since the dense search tokens have already been processed in the first stage, whereas the medium-density search tokens have not, we introduce a feature transformation module, as shown in Fig.~\ref{framework} (b), to align newly introduced shallow features with previously processed deeper features across sparsity levels.
The transformed medium-density search tokens are then concatenated with the existing sequence and fed into the second-stage Vision Transformer blocks for further feature learning.
\textbf{Finally}, the remaining search region with the last sparsity level, e.g., the sparse search region, is also processed by the feature transformation module and appended to the token sequence, yielding $F_{d,m,s} \in \mathbb{R}^{(N_z+3N_x)\times D}$. The resulting tokens are jointly fed into the third-stage Vision Transformer blocks for deeper feature learning.

By progressively incorporating search regions with different density levels through the proposed multi-stage learning strategy, our framework achieves two key advantages. First, it substantially reduces the computational burden caused by excessive search tokens, thereby improving computational efficiency. Second, the stage-wise integration facilitates progressive feature learning across different sparsity levels, enabling effective cross-scale interactions and enhancing the discriminability of the learned representations. Finally, the search features with different sparsity levels are separated from the token sequence and fused via element-wise summation to aggregate complementary information across all sparsity levels. The fused representation is then fed into the tracking head to predict the target location in the current frame.

\noindent $\bullet$ \textbf{Sparsity-aware Transformer Block}
As the grouped Transformer progressively introduces search regions with different sparsity levels, the feed-forward network (FFN) in each layer needs to process tokens with heterogeneous event-density distributions. These representations emphasize different visual cues. However, a shared FFN applies an identical transformation to all tokens, which may weaken sparsity-specific cues and suppress the complementary characteristics of sparse, medium-density, and dense event features, leading to suboptimal feature adaptation. To address this issue, inspired by the MoE paradigm~\cite{fedus2022switch}, we propose a sparsity-aware Transformer block that dynamically routes tokens to specialized experts according to their sparsity characteristics, thereby enhancing representation capacity and feature discriminability with limited additional computation.

We first revisit the standard FFN in a Transformer block, which performs token-wise nonlinear transformation and plays a crucial role in enhancing representation capacity. Each FFN is initialized from a pre-trained base model and consists of two linear layers, with the normalization layer omitted for simplicity. It can be parameterized as $[W_1,b_1,W_2,b_2]$, where $W_1 \in \mathbb{R}^{H \times D}$, $b_1 \in \mathbb{R}^{H \times 1}$, $W_2 \in \mathbb{R}^{D \times H}$, and $b_2 \in \mathbb{R}^{D \times 1}$. Here, $D$ denotes the token embedding dimension and $H$ denotes the hidden dimension. The FFN is formulated as
\begin{equation}
\mathrm{FFN}(x) = W_2 \left( \sigma \left( W_1 x + b_1 \right) \right) + b_2,
\label{ffn}
\end{equation}
where $x$ denotes the input tokens and $\sigma(\cdot)$ is the activation function.

To enable sparsity-aware specialization, as shown in Fig.~\ref{framework} (c), we decompose the original FFN into three sub-FFNs, each serving as a sparsity-specific expert for sparse, medium-density, or dense event representations. Meanwhile, the original full FFN is retained as a shared expert to preserve general representation capability. Together, these experts constitute the proposed sparsity-aware Mixture-of-Experts (SA-MoE) structure, as illustrated in Fig.~\ref{framework} (a).
Specifically, the first linear layer is partitioned along the hidden dimension into three sub-layers:
\begin{equation}
W_1 = \{ W_1^i \in \mathbb{R}^{(H/3)\times D} \mid i = 1,2,3 \},
\label{eq:w1}
\end{equation}
\begin{equation}
b_1 = \{ b_1^i \in \mathbb{R}^{(H/3)\times 1} \mid i = 1,2,3 \}.
\label{eq:b1}
\end{equation}
Similarly, the second linear layer is also partitioned along the hidden dimension:
\begin{equation}
W_2 = \{ W_2^i \in \mathbb{R}^{D \times (H/3)} \mid i = 1,2,3 \},
\label{eq:w2}
\end{equation}
while the output bias is evenly shared among experts:
\begin{equation}
b_2^* = b_2 / 3.
\label{eq:b2}
\end{equation}
Each expert $E_i$ is therefore defined as a lightweight sub-FFN, i.e., $E_i=[W_1^i,b_1^i,W_2^i,b_2^*]$. In this design, $\{E_1,E_2,E_3\}$ correspond to dense, medium-density, and sparse experts, respectively, while the original FFN is retained as a shared expert $E_s$.

\begin{figure*}
\center
\includegraphics[width=0.95\linewidth]{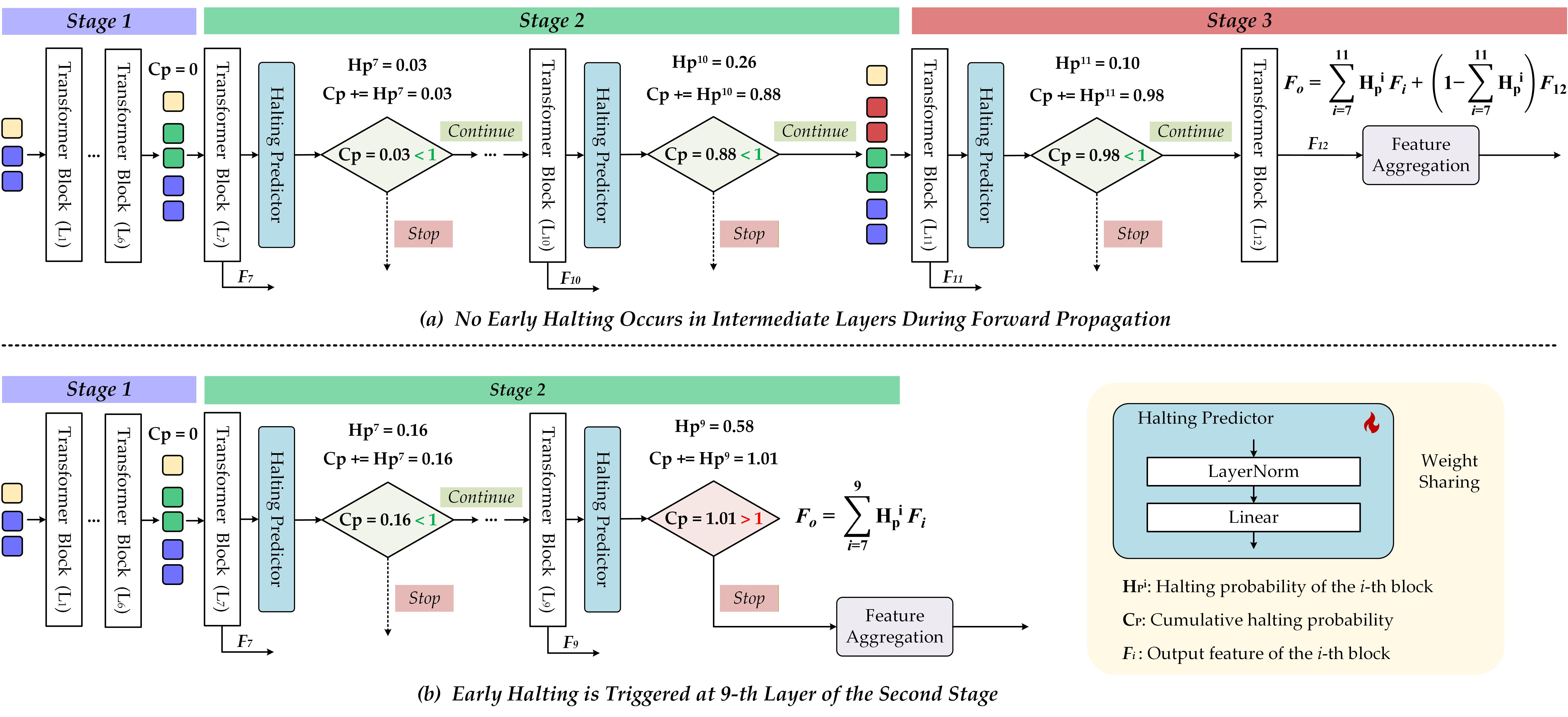}
\caption{The pipeline of the Dynamic Pondering Strategy (DPS). The upper part (a) shows that, in challenging scenarios, the model continues inference until the final layer to achieve higher accuracy before exiting. The lower part (b) illustrates that, in relatively simple scenarios, the model can perform effective tracking at intermediate layers with sufficient confidence. Under the regulation of DPS, the model adaptively terminates backbone inference at the 9-th layer.}
\label{Sub_Fig}
\end{figure*}

To dynamically select the most appropriate expert, we introduce a sparsity router. Specifically, we first apply global average pooling over the token dimension for both the template and all search region features, and then concatenate them along the channel dimension to obtain $R_{in} \in \mathbb{R}^{1 \times 2D}$. This representation is fed into an MLP-based router to predict the routing scores for three search regions with different sparsity levels:
\begin{equation}
S = \mathrm{MLP}(R_{in}),
\end{equation}
where $S \in \mathbb{R}^{1 \times 3}$ denotes the routing logits.
We then select the first $K$ dimensions as the active logits:
\begin{equation}
S_a = S_{1:K},
\end{equation}
where $K$ denotes the number of sparsity-level search regions currently introduced into the Transformer stage. To enable differentiable expert selection, we apply Gumbel-Softmax sampling to obtain the routing mask:
\begin{equation}
m = \mathrm{GumbelSoftmax}(S_a, \tau),
\end{equation}
where $\tau$ denotes the sampling temperature, which is set to $1$ in our implementation, and $m$ represents the resulting routing decision.
According to the routing result, the selected search features are assigned to the corresponding sparsity-specific expert. For example, if the medium-density search region is selected, its feature transformation is given by
\begin{equation}
X_m' = E_2(X_m).
\end{equation}
Finally, the output of the selected sparsity-specific expert is combined with the shared expert output via element-wise addition. If their dimensions are inconsistent, the sparsity-specific expert output is first aligned, e.g., by repetition, to match the dimensionality of the shared representation. The resulting representation is then used as the final output of the SA-MoE module. In practice, we replace the standard FFN with the proposed SA-MoE only in the first block of each stage, reducing redundant computation while preserving effective sparsity-aware feature adaptation.

\noindent $\bullet$ \textbf{Dynamic Pondering Strategy}
To further reduce the computational cost introduced by multi-sparsity search inputs, inspired by~\cite{yin2022vit, graves2016adaptive}, we propose a dynamic pondering strategy (DPS) that adaptively adjusts the inference depth of ViT layers. The overall pipeline is illustrated in Fig.~\ref{Sub_Fig}. To ensure sufficient feature learning in early layers, DPS is applied from the second stage, i.e., starting at layer 7.

We first initialize the cumulative halting probability $\mathbf{C_p}$ to zero. For the output feature of each Transformer block, a learnable halting predictor is employed to estimate the probability of terminating forward propagation at the current layer. Specifically, the halting predictor is implemented as a lightweight linear layer, which takes the globally average-pooled feature as input and outputs a scalar halting probability:
\begin{equation}
\mathbf{H}_p^i = \sigma\left( \mathrm{Linear}\left(\mathrm{GAP}(F_i)\right) \right),
\end{equation}
where $F_i$ denotes the output feature of the $i$-th Transformer block, $\mathbf{H}_p^i$ represents the corresponding halting probability, $\mathrm{GAP}(\cdot)$ denotes global average pooling, and $\sigma(\cdot)$ denotes the sigmoid activation function.

At each layer, the predicted halting probability is accumulated into the cumulative halting probability $\mathbf{C_p}$. If the updated $\mathbf{C_p} < 1$, the computation proceeds to the next block; otherwise, once $\mathbf{C_p} \geq 1$, forward propagation is terminated. The final representation is obtained by aggregating features from multiple layers:
\begin{equation}
F_o =
\begin{cases}
\sum_{i=7}^{11} \mathbf{{H_p}^i} F_i + \left(1 - \sum_{i=7}^{11} \mathbf{{H_p}^i}\right) F_{12}, & \text{if } L = 12, \\
\sum_{i=7}^{L} \mathbf{{H_p}^i} F_i, & \text{otherwise},
\end{cases}
\end{equation}
where $F_i$ denotes the output feature of the $i$-th block, $F_o$ represents the final aggregated feature, and $L$ denotes the halting layer. At the final layer, the halting probability is defined as the remaining probability mass after subtracting the accumulated probabilities of all preceding layers from $1$, ensuring that the aggregation weights across layers sum to $1$.

During training, we introduce a dynamic pondering loss $\mathcal{L}_{\text{ponder}}$ to penalize the number of executed layers after DPS is activated. Since DPS starts from the second stage, i.e., the 7-th layer, the loss is defined as
\begin{equation}
\mathcal{L}_{\text{ponder}} = L - 6,
\end{equation}
where $L$ denotes the halting layer. This loss is incorporated into the overall tracking objective, encouraging the model to maintain robust tracking performance with fewer executed layers. During inference, DPS adaptively adjusts the inference depth according to scenario complexity and tracking difficulty, allowing computation to terminate at an appropriate layer.

\subsection{Tracking Head and Loss Function}
Following OSTrack~\cite{ye2022ostrack}, we adopt a center-based tracking head. Specifically, the search-region features are first reshaped into a feature map and then processed by the convolution–batch normalization–ReLU (Conv-BN-ReLU) layers to generate four key components: (1) a target classification score map that predicts the likelihood of the target at each spatial location; (2) local offsets for refining the bounding box center coordinates; (3) normalized bounding box sizes (width and height); and (4) the final predicted bounding boxes, representing the estimated target positions in the current frame.

During training, we follow the loss formulation of OSTrack~\cite{ye2022ostrack}, which includes three complementary objectives: Focal Loss ($\mathcal{L}_{focal}$) for target classification, L1 Loss ($\mathcal{L}_1$) for offset regression, and GIoU Loss ($\mathcal{L}_{GIoU}$) for bounding box overlap optimization.
In addition, to optimize the proposed dynamic pondering strategy, we introduce a pondering loss $\mathcal{L}_{\mathrm{ponder}}$ that penalizes the number of activated ViT layers during forward propagation. The overall training objective is defined as
\begin{align}
    \label{lossFunction} 
    & \mathcal{L}_{total} = \lambda_1 \mathcal{L}_{focal} + \lambda_2 \mathcal{L}_{1} + \lambda_3 \mathcal{L}_{GIoU} + \lambda_4 \mathcal{L}_{ponder}.
\end{align}
Here, $\lambda_1$, $\lambda_2$, and $\lambda_3$ are weighting coefficients for the tracking losses and are set to $1$, $5$, and $2$, respectively. For $\lambda_4$, we adopt a cosine-based scheduling strategy:
\[
\lambda_4 = \alpha \cdot \left(1 - \cos\left(\pi \cdot R\right)\right),
\]
where $\alpha$ is a hyperparameter controlling the maximum weight of the pondering loss. 
The scheduling ratio $R$ is defined as
\begin{equation}
R =
\begin{cases}
0, & e < e_{\mathrm{start}}, \\
\dfrac{e-e_{\mathrm{start}}}{e_{\mathrm{total}}-e_{\mathrm{start}}}, & e \geq e_{\mathrm{start}},
\end{cases}
\end{equation}
where $e$, $e_{\mathrm{start}}$, and $e_{\mathrm{total}}$ denote the current training epoch, the activation epoch of the dynamic pondering strategy, and the total number of training epochs, respectively. This scheduling strategy gradually increases the contribution of the pondering loss, enabling stable optimization of DPS while encouraging efficient inference.

\begin{table*}
\center
\small     
\caption{Experimental results (SR/PR) on FE240hz dataset.} 
\label{FE240table}
\resizebox{0.88\textwidth}{!}{
\begin{tabular}{ccccccccccc}
\hline 
\textbf{STNet~\cite{zhang2022spiking}}  &\textbf{TransT~\cite{chen2021transt}}  &\textbf{STARK~\cite{yan2021stark}}        &\textbf{SiamFC++~\cite{xu2020siamfc++}}  &\textbf{DiMP~\cite{goutam2019Dimp}}  &\textbf{ATOM~\cite{danelljan2019atom}}  &\textbf{OSTrack~\cite{ye2022ostrack}} \\ 
58.5/89.6      &56.7/89.0        &55.4/83.7    &54.5/85.3   &53.4/88.2  &52.8/80.0     & 57.1/89.3 \\ 
\hline 
\textbf{MixFormer~\cite{cui2022mixformer}}   &\textbf{AQATrack~\cite{xie2024autoregressive}}   &\textbf{HDETrack~\cite{wang2024event}}  &\textbf{MambaEvT~\cite{mambaevt2025}} &\textbf{TSETrack~\cite{zhang2026efficient}}  &\textbf{Ours-I}   &\textbf{Ours-II} \\ 
57.1/88.6    &59.9/92.6        &59.8/92.2     &58.1/92.0     &59.5/90.9    &\textbf{61.9/93.5}   &61.3/93.0  \\         
\hline 
\end{tabular}
}
\end{table*}

\section{Experiments} 

\subsection{Datasets and Evaluation Metric}

To comprehensively evaluate the effectiveness of the proposed PSMTrack, we conduct extensive experiments on three public event-based tracking datasets, including \textbf{FE240hz}~\cite{zhang2021fe108}, \textbf{COESOT}~\cite{tang2025revisiting}, and \textbf{EventVOT}~\cite{wang2024event}. 
For {FE240hz}~\cite{zhang2021fe108} and {COESOT}~\cite{tang2025revisiting}, we compare PSMTrack with other methods using only unimodal event data as input. Each tracker is retrained and evaluated on the corresponding event frames following the original parameter settings of the respective method. A brief introduction to these event-based tracking datasets is provided below.

$\bullet$ \textbf{FE240hz dataset}: FE240hz was collected using a grayscale DVS346 event camera and contains 71 training videos and 25 testing videos. It provides more than 1.13 million annotations over 143K image frames and their corresponding event data. The dataset covers various challenging tracking scenarios, such as motion blur and high dynamic range conditions. More details can be found on the GitHub\footnote{\url{https://github.com/Jee-King/ICCV2021_Event_Frame_Tracking?tab=readme-ov-file}}.

$\bullet$ \textbf{COESOT dataset}: COESOT is a large-scale, category-wide RGB-event tracking dataset covering 90 object categories and 1,354 video sequences, with a total of 478,721 event frames. To systematically evaluate tracking robustness, the dataset defines 17 challenging factors. It is split into 827 training videos and 527 testing videos, providing a comprehensive benchmark for event-based tracking research. More details can be found on the GitHub\footnote{\url{https://github.com/Event-AHU/COESOT}}.

$\bullet$ \textbf{EventVOT dataset}: EventVOT is the first large-scale, high-resolution event-based tracking dataset captured using a Prophesee camera. It contains 841 training videos and 282 testing videos, covering 19 object categories, such as UAVs, basketball, and pedestrians. Each video is uniformly divided into 499 frames to facilitate high-quality annotation. The dataset further defines 14 challenging attributes, including fast motion and small objects, to comprehensively evaluate tracking robustness. More details can be found on the GitHub\footnote{\url{https://github.com/Event-AHU/EventVOT_Benchmark}}.

For evaluation, we adopt three widely used tracking metrics: \textbf{Success Rate (SR)}, \textbf{Precision Rate (PR)}, and \textbf{Normalized Precision Rate (NPR)}. Specifically, SR is defined as the proportion of frames in which the Intersection over Union (IoU) between the predicted and ground-truth bounding boxes exceeds a given threshold. PR measures the percentage of frames where the Euclidean distance between the predicted and ground-truth target centers is below a predefined threshold, which is set to 20 pixels by default. NPR further normalizes the center-location error by the diagonal length of the ground-truth bounding box to account for scale variations. In addition, we report \textbf{Frames Per Second (FPS)} to evaluate tracking efficiency.

\subsection{Implementation Details} 
Our framework is built upon OSTrack~\cite{ye2022ostrack}, with ViT-B adopted as the backbone network. The backbone is divided into three stages with 6, 4, and 2 layers, respectively, and is initialized with pretrained weights~\cite{ye2022ostrack}. The template and search regions are cropped and resized to $128 \times 128$ and $256 \times 256$, respectively. During training, grayscale conversion is applied as data augmentation with a probability of $5\%$.
The model is trained for 50 epochs with a batch size of 32 using the AdamW optimizer~\cite{loshchilov2018adamw}. The initial learning rate and weight decay are set to $4 \times 10^{-4}$ and $1 \times 10^{-4}$, respectively. A step learning rate scheduler is employed, where the learning rate is decayed by a factor of 0.1 at epoch 40. For the dynamic pondering strategy, the activation epoch is set to 10. The hyperparameter $\alpha$ is set to 0.04, 0.06, and 0.10 for EventVOT, COESOT, and FE240hz, respectively.
Our code is implemented in Python based on PyTorch~\cite{paszke2019pytorch}, and all experiments are conducted on a server equipped with an NVIDIA RTX 4090 GPU. More implementation details will be made available in our GitHub repository.

\subsection{Comparison on Public Benchmark Datasets}

\begin{table}
\center
\small   
\caption{Overall tracking performance on the COESOT dataset. } 
\label{COESOT_results}
\resizebox{0.95\columnwidth}{!}{ 
\begin{tabular}{l|l|c|cccc}
\hline 
\textbf{No.} & \textbf{Trackers} & \textbf{Source}   & \textbf{SR}  &\textbf{PR}   &\textbf{NPR} \\
\hline
01 &\textbf{TrDiMP~\cite{wang2021TrDiMP}}   & CVPR21     &50.7       &59.2      &58.4           \\ 
02 &\textbf{ToMP50~\cite{mayer2022Tomp}}   &  CVPR22   &46.3       &55.2      &56.0           \\ 
03 &\textbf{OSTrack~\cite{ye2022ostrack}}   &  ECCV22   &50.9       &61.8       &61.5          \\ 
04 &\textbf{AiATrack~\cite{gao2022AIa}}   &ECCV22   &50.6       &59.5       &59.2           \\ 
05 &\textbf{STARK~\cite{yan2021stark}}   &  ICCV21    &40.8      &44.5      &46.1           \\ 
06 &\textbf{TransT~\cite{chen2021transt}}   &  CVPR21     &45.6       &54.3       &54.2           \\ 
07 &\textbf{DiMP50~\cite{goutam2019Dimp}}  &  ICCV19     &53.8      &64.8       &65.1           \\ 
08 &\textbf{PrDiMP~\cite{martin2020PrDimp}}  &  CVPR20     &47.5       &57.8      &57.9           \\ 
09 &\textbf{KYS~\cite{bhat2022SKys}}   &   ECCV20      &42.6       &52.7       &52.1           \\ 
10 &\textbf{MixFormer~\cite{cui2022mixformer}}   & CVPR22   &44.4     &50.2      &51.1           \\ 
11 &\textbf{ATOM~\cite{danelljan2019atom}}   & CVPR19    &42.1      &50.4        &51.3          \\ 
12 &\textbf{SimTrack~\cite{chen2022simtrack}}   & ECCV22  & 48.3      &55.7       &56.6           \\  
13 &\textbf{HDETrack~\cite{wang2024event}}    &CVPR24            &53.1       &64.1      &64.5          \\ 
14 &\textbf{MCITrack~\cite{kang2025exploring}}    &AAAI25            &56.3       & 67.6     & 67.3       \\ 
15 &\textbf{LMTrack~\cite{xu2025less}}    &AAAI25            &56.4       &\textbf{69.1}      &67.9         \\ 
26 &\textbf{UTPTrack~\cite{wu2026utptrack}}    &CVPR26  &55.2  &67.1  &66.9    \\ 
17 &\textbf{SpikeTrack~\cite{zhang2026spiketrack}}    &CVPR26            &54.2       &66.0      &65.9         \\ 
\hline
18 &\textbf{Ours-I} & - &\textbf{56.6}   &69.0   &\textbf{68.7} \\
19 &\textbf{Ours-II} & - &56.2   &68.5   &68.1 \\
\hline
\end{tabular}
}
\end{table}

\noindent $\bullet$ \textbf{Results on FE240hz Dataset.~} 
As shown in Table~\ref{FE240table}, the proposed method achieves new state-of-the-art (SOTA) performance on the FE240hz dataset. Specifically, Ours-I denotes the proposed method without DPS, achieving SR and PR scores of 61.9\% and 93.5\%, respectively, while Ours-II denotes the method equipped with DPS, achieving SR and PR scores of 61.3\% and 93.0\%, respectively. Compared with recent SOTA trackers, such as HDETrack and TSETrack, Ours-I improves SR by 1.5\% and 2.4\%, respectively, demonstrating the effectiveness and robustness of the proposed tracking framework. With the dynamic pondering strategy, Ours-II enables more efficient tracking while maintaining competitive accuracy.

\begin{table}
\centering
\small   
\caption{Overall Tracking Performance on EventVOT Dataset. } 
\label{EventVOTtable}
\resizebox{0.95\columnwidth}{!}{ 
\begin{tabular}{l|l|c|lll|l}
\hline \toprule [0.5 pt]
\textbf{No.} & \textbf{Trackers} & \textbf{Source}   & \textbf{SR}  &\textbf{PR}   &\textbf{NPR}   &\textbf{FPS}\\
\hline
01    &  \textbf{DiMP50~\cite{goutam2019Dimp}}  &  ICCV19       &52.6   &51.1   &67.2    &43  \\
02    &  \textbf{ATOM~\cite{martin2019Atom}}   & CVPR19     &44.4   &44.0   &57.5      &30  \\
03    &  \textbf{PrDiMP~\cite{martin2020PrDimp}}  &  CVPR20       &55.5   &57.2   &70.4      &30  \\
04    &  \textbf{KYS~\cite{bhat2022SKys}}   &   ECCV20         &38.7   &37.3   &49.8     &20  \\ 
05    &  \textbf{TrDiMP~\cite{wang2021TrDiMP}} & CVPR21     &39.9   &34.8   &48.7     &26   \\ 
06    &  \textbf{STARK~\cite{yan2021stark}}   &  ICCV21     &44.5   &39.6  &55.7      &42  \\ 
07    &  \textbf{TransT~\cite{chen2021transt}}   &  CVPR21     &54.3  &56.5  &68.8      &50  \\ 
08    &  \textbf{ToMP50~\cite{mayer2022Tomp}}   &  CVPR22   &37.6   &32.8   &47.4      &25  \\ 
09    &  \textbf{OSTrack~\cite{ye2022ostrack}}   &  ECCV22   &55.4  &60.4   &71.1      &105  \\
10    &  \textbf{AiATrack~\cite{gao2022AIa}}   &  ECCV22     &57.4   &59.7   &72.8      &38  \\ 
11    &  \textbf{MixFormer~\cite{cui2022mixformer}}   & CVPR22     &49.9   &49.6   &63.0      &25  \\
12    &  \textbf{SimTrack~\cite{chen2022simtrack}}   & ECCV22     &55.4   &57.5  &69.9      &40  \\ 
13    &  \textbf{ROMTrack~\cite{cai2023robust}} &ICCV23  &56.3  &60.2 &71.8    &62 \\
14    &  \textbf{ARTrack~\cite{wei2023autoregressive}} &CVPR23  &52.0  &52.0 &65.6    &26  \\ 
15    &  \textbf{ODTrack~\cite{zheng2024odtrack}}      &AAAI24  &56.8   &62.2  &72.1      &32  \\ 
16    &  \textbf{EVPTrack~\cite{shi2024explicit}}      &AAAI24  &58.4   &62.8  &73.9        &71 \\
17    &  \textbf{ARTrackV2~\cite{bai2024artrackv2}} &CVPR24   &55.4 &56.6 &70.0      &94  \\
18    &  \textbf{AQATrack~\cite{xie2024autoregressive}}      &CVPR24  &59.2   &65.0  &74.3     &68 \\
19    &  \textbf{HDETrack~\cite{wang2024event}}    &CVPR24 &57.8   &62.2  &73.5    &105 \\ 
20    &  \textbf{MambaEvT~\cite{mambaevt2025}}    &TCSVT25 &56.5   &56.7  &65.5       &28 \\ 
21    &  \textbf{MCITrack~\cite{kang2025exploring}}    &AAAI25  &59.9   &66.1  &74.6    & 35        \\
22    &  \textbf{LMTrack~\cite{xu2025less}}    &AAAI25  &57.3       &63.3      &71.2    & 56        \\ 
23    &  \textbf{UTPTrack~\cite{wu2026utptrack}}    &CVPR26  &59.7   &66.2  &75.2   & 95 \\ 
24   &  \textbf{SpikeTrack~\cite{zhang2026spiketrack}}    &CVPR26  &58.6   & 63.8 & 74.2   & 72 \\ 
\hline
25   &  \textbf{Ours-I}      &-   & \textbf{60.3}     & \textbf{66.6}   & \textbf{75.6}   &56   \\ 
26   &  \textbf{Ours-II}      &-  &59.9  &66.2  &75.4   &79   \\ 
\hline \toprule [0.5 pt]
\end{tabular}
}
\end{table}

\begin{figure}
\center
\includegraphics[width=0.45\textwidth]{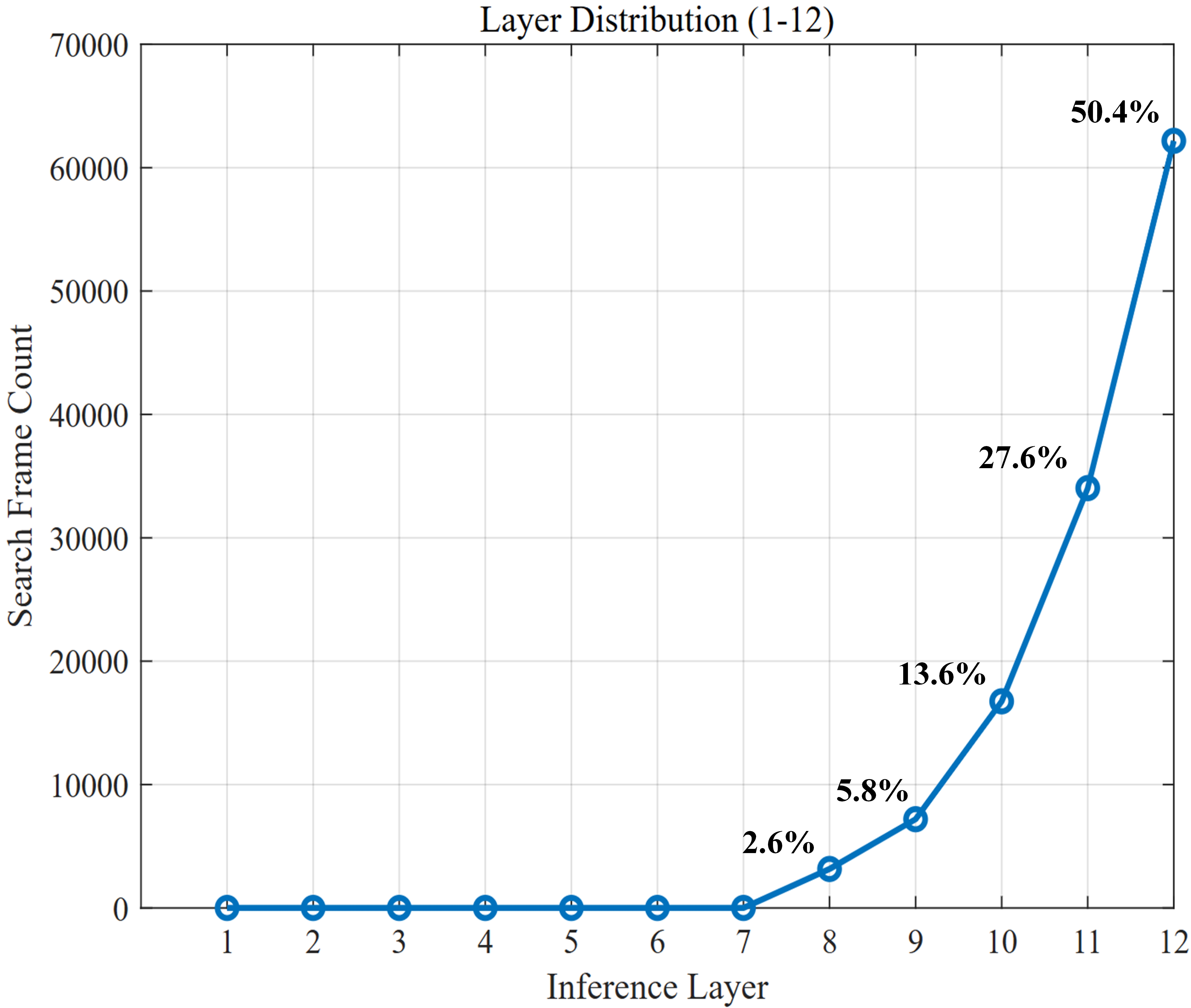}
\caption{Inference layer statistics on the EventVOT dataset.}  
\label{layer_distribution}
\end{figure}

\begin{table*}
\center
\small     
\caption{Component Analysis results on the EventVOT dataset.} 
\label{CAResults} 
\begin{tabular}{c|ccccc|ccc|cc} 		
\hline 
\textbf{No.} & \textbf{Baseline}  &\textbf{Multi-Sparsity} &\textbf{Multi-Stage} &\textbf{SA-MoE}   &\textbf{DPS}    &\textbf{SR}   &\textbf{PR}   & \textbf{NPR} &\textbf{Params} &\textbf{FPS}   \\
\hline 
1 &\cmark  &          &         &         &         &58.5   &63.5   &73.6   &92.5M  &105   \\
2 &\cmark  &\cmark    &         &         &         &59.2   &65.3   &74.2   &92.5M  & 52  \\
3 &\cmark  &\cmark    &\cmark   &         &         &59.6   &65.6   &74.9   &93.1M  & 64   \\
4 &\cmark  &\cmark    &\cmark   &\cmark   &         & \textbf{60.3}     & \textbf{66.6}   & \textbf{75.6}   &108.5M  & 56 \\
5 &\cmark  &\cmark    &\cmark   &\cmark   &\cmark   &59.9   &66.2   &75.4   &108.5M   & 79  \\
\hline
\end{tabular}
\end{table*}

\noindent $\bullet$ \textbf{Results on COESOT Dataset.~}
As shown in Table~\ref{COESOT_results}, we further report the tracking results on the large-scale COESOT benchmark. Specifically, Ours-I denotes the proposed method without DPS, which outperforms all compared methods in terms of SR and NPR, while showing only a marginal gap of 0.1\% in PR compared with the best-performing tracker. These results demonstrate the effectiveness of the proposed framework in exploiting multi-sparsity event representations for robust tracking. Ours-II denotes the method equipped with DPS, which incurs only a slight average accuracy drop of 0.5\% compared with Ours-I while improving tracking efficiency.

\noindent $\bullet$ \textbf{Results on EventVOT Dataset.~}
As shown in Table~\ref{EventVOTtable}, we report the tracking results on the high-resolution event-based tracking dataset EventVOT. Specifically, Ours-I denotes the proposed method without DPS, which outperforms all compared SOTA trackers and achieves SR, PR, and NPR scores of 60.3\%, 66.6\%, and 75.6\%, respectively. Ours-II denotes the method equipped with DPS, which incurs only a slight average accuracy drop of approximately 0.3\% compared with Ours-I while significantly improving the tracking speed from 56 FPS to 79 FPS.
Moreover, with DPS, Ours-II still maintains a clear accuracy advantage over existing SOTA methods. For example, compared with SpikeTrack, Ours-II improves SR, PR, and NPR by 1.3\%, 2.4\%, and 1.2\%, respectively. Compared with MCITrack, Ours-II achieves comparable tracking accuracy while being more than twice as fast. Meanwhile, we further analyze the distribution of halting layers under the dynamic pondering strategy. As shown in Fig.~\ref{layer_distribution}, the model terminates inference early in approximately 50\% of the scenarios. These results demonstrate the superiority of the proposed method for event-based tracking and its favorable trade-off between tracking accuracy and efficiency.

\begin{table}
\center
\small     
\caption{Ablation Studies on the EventVOT dataset.} 
\label{tab: Ablation_Studies}
\resizebox{0.98\columnwidth}{!}{ 
\begin{tabular}{l|llll}
\rowcolor{gray!20}
\hline \toprule [0.5 pt] 
\textbf{\# Event Representation}    &\textbf{SR}   & \textbf{PR}  & \textbf{NPR} \\
\hline
\text{1. Event Time Surface}     &54.9     &57.9    &69.8    \\
\text{2. Event Reconstruction Images}     &57.5  &63.9   &71.5 \\
\text{3. Event Voxels}      & 12.6   &18.7    &21.2  \\
\text{4. Event Frames (Sparse)}      & 56.9   &61.1    &72.3  \\
\text{5. Event Frames (Medium)}      & 58.5   &63.5    &73.6  \\
\textbf{6. Event Frames (Dense)}      & \textbf{58.7}   & \textbf{64.8}    &\textbf{74.1}  \\
\hline 
\rowcolor{gray!20}
\textbf{\# Multi-sparsity Input Order} & \textbf{SR} & \textbf{PR} & \textbf{NPR} \\
\hline
\text{1. Sparse → Medium → Dense} &57.5  &61.9  &72.8  \\
\text{2. Dense → Sparse → Medium} & 59.2  & 65.3  & 74.6 \\
\text{3. Medium → Sparse → Dense} &59.0  &64.5  &74.4  \\
\textbf{4. Dense → Medium → Sparse (Ours)} & \textbf{59.6} & \textbf{65.6} & \textbf{74.9} \\
\hline
\rowcolor{gray!20}
\textbf{\# Layer Allocation (Dense / Mid / Sparse) }    &\textbf{SR}   & \textbf{PR}  & \textbf{NPR}  \\
\hline
\text{1. 4 / 4 / 4 Layers }    &59.4  &65.4  &74.5  \\ 
\text{2. 6 / 3 / 3 Layers }    & 59.6  & 65.2  & 74.6 \\
\text{3. 2 / 4 / 6 Layers }    &59.7  &65.4  &74.9  \\ 
\textbf{4. 6 / 4 / 2 Layers  (Ours)}     & \textbf{59.6} & \textbf{65.6} & \textbf{74.9}  \\ 
\text{5. Ours \textit{w/o} Feature Transformation} &59.3  &65.4 &74.4 \\
\hline 
\rowcolor{gray!20}
\textbf{\# Sparsity Experts}    &\textbf{SR}   & \textbf{PR}  & \textbf{NPR}  \\
\hline
\text{1. \textit{w/o} Sparsity Experts }                   & 59.6    & 65.6   & 74.9   \\
\text{2. \textit{w/o} Shared Experts }                   &59.4     &65.3  &74.3     \\
\text{3. Single Sparsity Expert }                   & 59.4    &66.1  &74.6     \\ 
\textbf{4. Multi-sparsity Experts (Ours)  }        & \textbf{60.3}     & \textbf{66.6}   & \textbf{75.6}  \\ 
\hline
\rowcolor{gray!20}
\textbf{\# SA-MoE Layers}    &\textbf{SR}   & \textbf{PR}  & \textbf{NPR}  \\
\hline
\text{1. \textit{w/o} SA-MoE }                   & 59.6    & 65.6   & 74.9     \\
\text{2. All Layers }                   &59.7     &66.1  &74.9     \\ 
\text{3. Only Stage-2 Layers  }        &59.9     &66.2      &75.2  \\ 
\text{4. Only Stage-3 Layers  }        &59.4    &65.8      &74.5  \\ 
\textbf{5. First Layer of Each Stage (Ours)}          & \textbf{60.3}     & \textbf{66.6}   & \textbf{75.6} \\
\hline
\rowcolor{gray!20}
\textbf{\# Depth Compression Methods}    &\textbf{SR}   & \textbf{PR}  & \textbf{NPR}  \\
\hline
\text{1. Stochastic Layer Exit }                   &58.0     &63.2  &73.6     \\ 
\text{2. CompressTracker~\cite{hong2025general} }                   &54.6      &58.3  &70.1     \\
\text{3. Similarity-Guided Layer-Adaptive~\cite{xue2025similarity} }                   &56.1     &59.3  &71.0     \\
\textbf{4. Dynamic Pondering Strategy (Ours)  }        & \textbf{59.9}     & \textbf{66.2}   & \textbf{75.4}  \\ 
\hline \toprule [0.5 pt] 
\end{tabular}
}
\end{table}

\subsection{Ablation Study}

\noindent $\bullet$ \textbf{Component Analysis.~}
To verify the effectiveness of each proposed component, we conduct a comprehensive ablation study on the EventVOT dataset, with the results reported in Table~\ref{CAResults}. The baseline model, shown in the first row, achieves SR, PR, and NPR scores of 58.5\%, 63.5\%, and 73.6\%, respectively, with 92.5M parameters and a running speed of 105 FPS. Based on this baseline, we progressively introduce each component to evaluate its contribution.
Specifically, in the second row, we introduce multi-sparsity event representations and directly concatenate them as inputs to the backbone network. This strategy brings noticeable accuracy improvements, but reduces efficiency due to the increased computational cost of processing additional tokens. 
In the third row, we design a multi-stage backbone that progressively incorporates event representations with different sparsity levels, thereby alleviating the computational burden. This design further improves tracking accuracy while enhancing efficiency.
In the fourth row, we incorporate the sparsity-aware MoE module, which encourages expert specialization across different sparsity levels and leads to additional accuracy gains. 
Finally, in the last row, we introduce the dynamic pondering strategy, which increases the inference speed to 79 FPS with only a marginal accuracy drop, ultimately achieving a more favorable trade-off between tracking accuracy and efficiency.

\noindent $\bullet$ \textbf{Analysis on Different Event Representation.~}
To investigate the influence of different event representations on tracking performance, we conduct an ablation study on the EventVOT dataset by comparing four representative forms: event time surface, event reconstructed image, event voxel, and event frame. The event frame is further categorized into sparse, medium-density, and dense representations. As shown in Table~\ref{tab: Ablation_Studies}, the choice of event representation significantly affects tracking accuracy. Among all representations, event voxel yields the worst performance, which may be attributed to information loss during voxelization that weakens its spatial modeling capability. In contrast, the dense event frame achieves the best performance, benefiting from its higher event density, which provides richer motion cues.

\noindent $\bullet$ \textbf{Analysis on Multi-sparsity Input Order.~}
We further analyze the effect of different input orders of sparse, medium-density, and dense representations on tracking accuracy. As shown in Table~\ref{tab: Ablation_Studies}, the Dense $\rightarrow$ Medium-density $\rightarrow$ Sparse order, i.e., Ours, achieves the best performance and outperforms all other strategies. Compared with the Sparse $\rightarrow$ Medium-density $\rightarrow$ Dense order, which achieves 57.5\% in SR, and the Medium-density $\rightarrow$ Sparse $\rightarrow$ Dense order, which achieves 59.0\% in SR, our strategy first leverages dense representations to capture rich motion cues, and then progressively incorporates medium-density and sparse representations to refine structural and contour details. This progressive design facilitates effective utilization of complementary event cues and enhances overall tracking stability.

\noindent $\bullet$ \textbf{Analysis on Layer Allocation (Dense / Medium-density / Sparse).~}
To investigate the trade-off between tracking performance and computational complexity, we compare four layer allocation strategies for dense, medium-density, and sparse feature processing, i.e., 4/4/4, 6/3/3, 2/4/6, and 6/4/2 (Ours). As shown in Table~\ref{tab: Ablation_Studies}, the 6/4/2 configuration achieves SR and PR scores of 59.6\% and 65.6\%, respectively. It delivers performance comparable to the 2/4/6 setting while requiring lower computational cost, owing to a more efficient allocation of layer resources. In addition, we further ablate the feature transformation module and observe an accuracy drop after removing it, demonstrating its effectiveness in aligning deep and shallow features.

\noindent $\bullet$ \textbf{Analysis on Sparsity Experts.~}
We further investigate the impact of sparsity experts on model performance by comparing four configurations: without sparsity experts, without the shared expert, with a single sparsity expert, and with the proposed multi-sparsity experts (Ours). The results show that the proposed multi-sparsity experts achieve the best overall performance, with SR, PR, and NPR scores of 60.3\%, 66.6\%, and 75.6\%, respectively. In particular, it outperform the single-sparsity-expert configuration by 1.0\% in NPR, further validating that the proposed SA-MoE module explicitly models event representations with different sparsity levels through multi-sparsity experts, thereby enhancing feature extraction and discriminability for improved tracking accuracy.

\noindent $\bullet$ \textbf{Analysis of SA-MoE Layer Placement.~}
To determine the optimal layer placement of SA-MoE within the ViT backbone, we compare four configurations:without SA-MoE, applying SA-MoE to all layers, applying SA-MoE only to Stage-2 layers, and applying SA-MoE to the first layer of each stage (Ours). The results show that the proposed placement achieves the best overall performance. Specifically, since the first layer of each stage is responsible for adapting newly introduced sparsity-level features, applying SA-MoE only at these layers enables effective sparsity-aware feature adaptation while avoiding redundant routing across the entire backbone.

\noindent $\bullet$ \textbf{Analysis on Depth Compression Methods.~}
Furthermore, we compare the proposed dynamic pondering strategy (DPS) with other model depth compression approaches. Specifically, we first evaluate Stochastic Layer Exit and observe a noticeable accuracy degradation, mainly due to the stochastic nature of layer skipping, which may introduce instability during inference. In addition, CompressTracker and Similarity-Guided Layer-Adaptive, two representative depth compression methods for visual tracking, also yield substantially lower accuracy than our approach under their respective settings. These results demonstrate that DPS achieves a more favorable trade-off between tracking accuracy and efficiency.

\begin{figure}
\includegraphics[width=\linewidth]{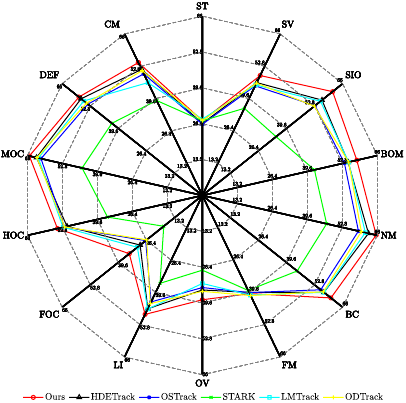}
\caption{Tracking results (SR) under each challenging factor.} 
\label{attributeResults}
\end{figure}

\noindent $\bullet$ \textbf{Results Under Each Attribute.~}
As shown in Fig.~\ref{attributeResults}, we further report the performance of the proposed method and other state-of-the-art trackers under different challenging attributes. Benefiting from the proposed multi-sparsity relation modeling strategy, our method achieves leading performance on most attributes. In particular, it shows clear advantages under SIO (similar interfering objects), NM (no motion), BC (background clutter), and FOC (full occlusion), further demonstrating the robustness of the proposed tracker against diverse real-world challenges.

\begin{figure*}
\center
\includegraphics[width=\textwidth]{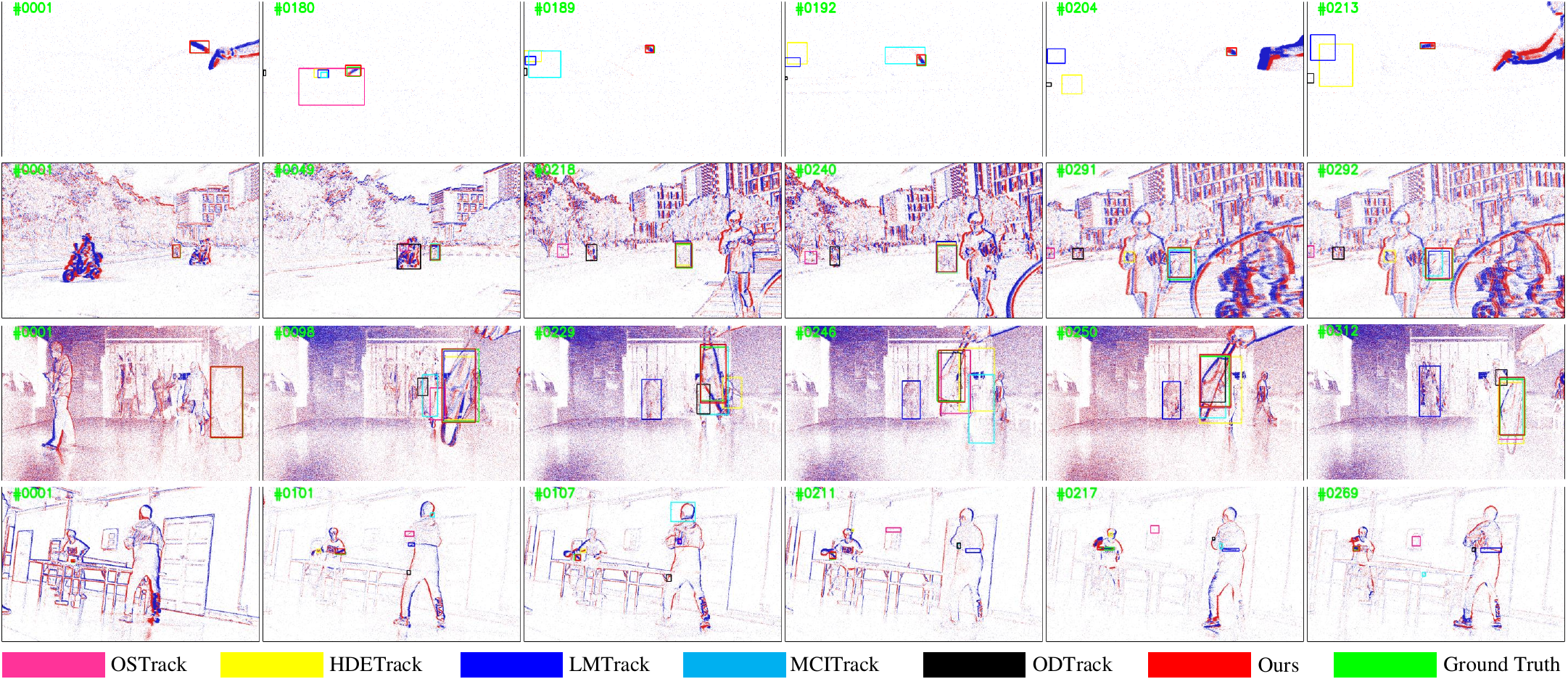}
\caption{Visualization of the tracking results of our method and other SOTA trackers.}  
\label{trackingResults}
\end{figure*}

\begin{figure}
\center
\includegraphics[width=0.98\linewidth]{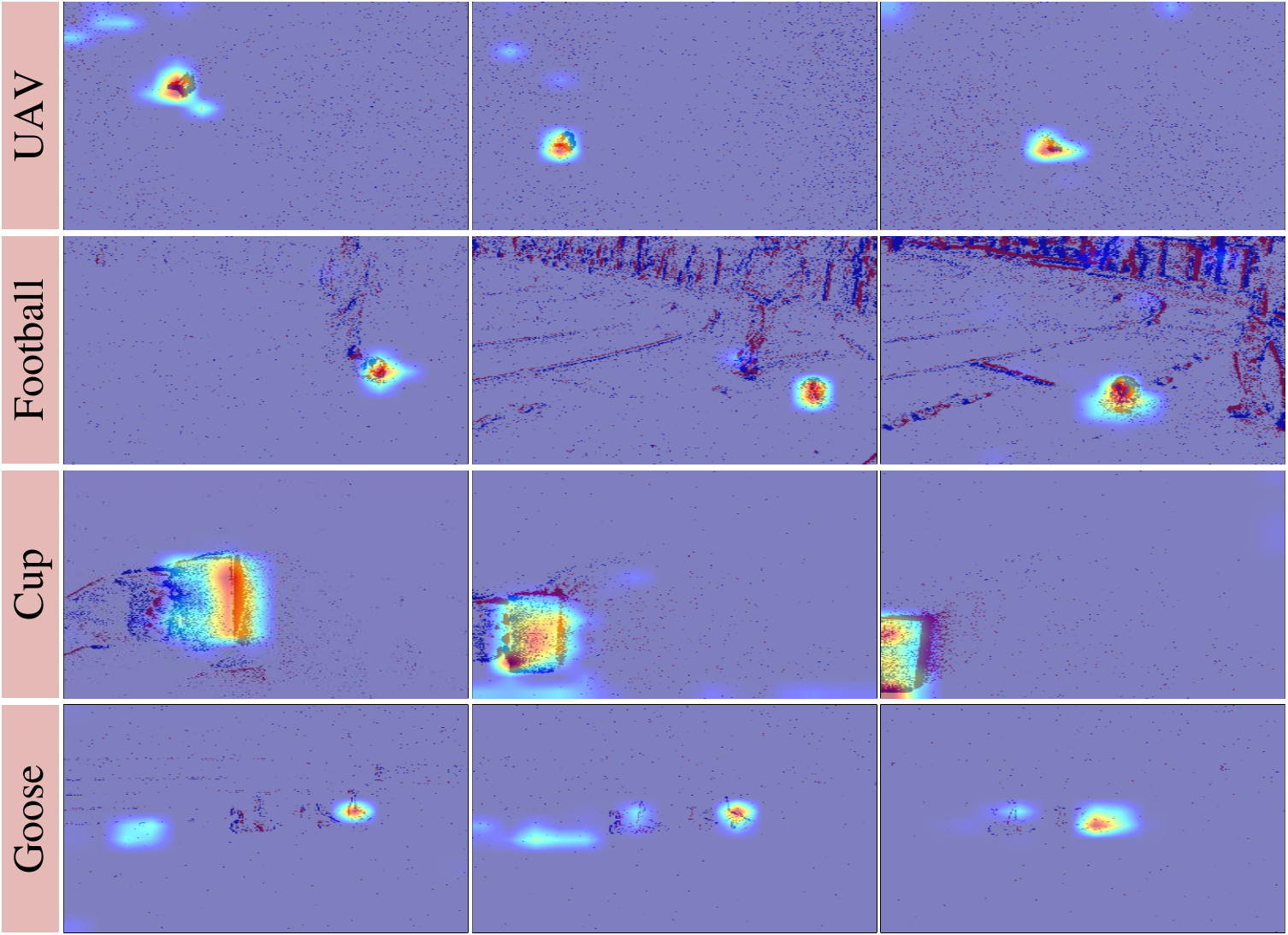}
\caption{Attention maps predicted by our method.}  
\label{attention_map}
\end{figure} 

\begin{figure}
\center
\includegraphics[width=0.98\linewidth]{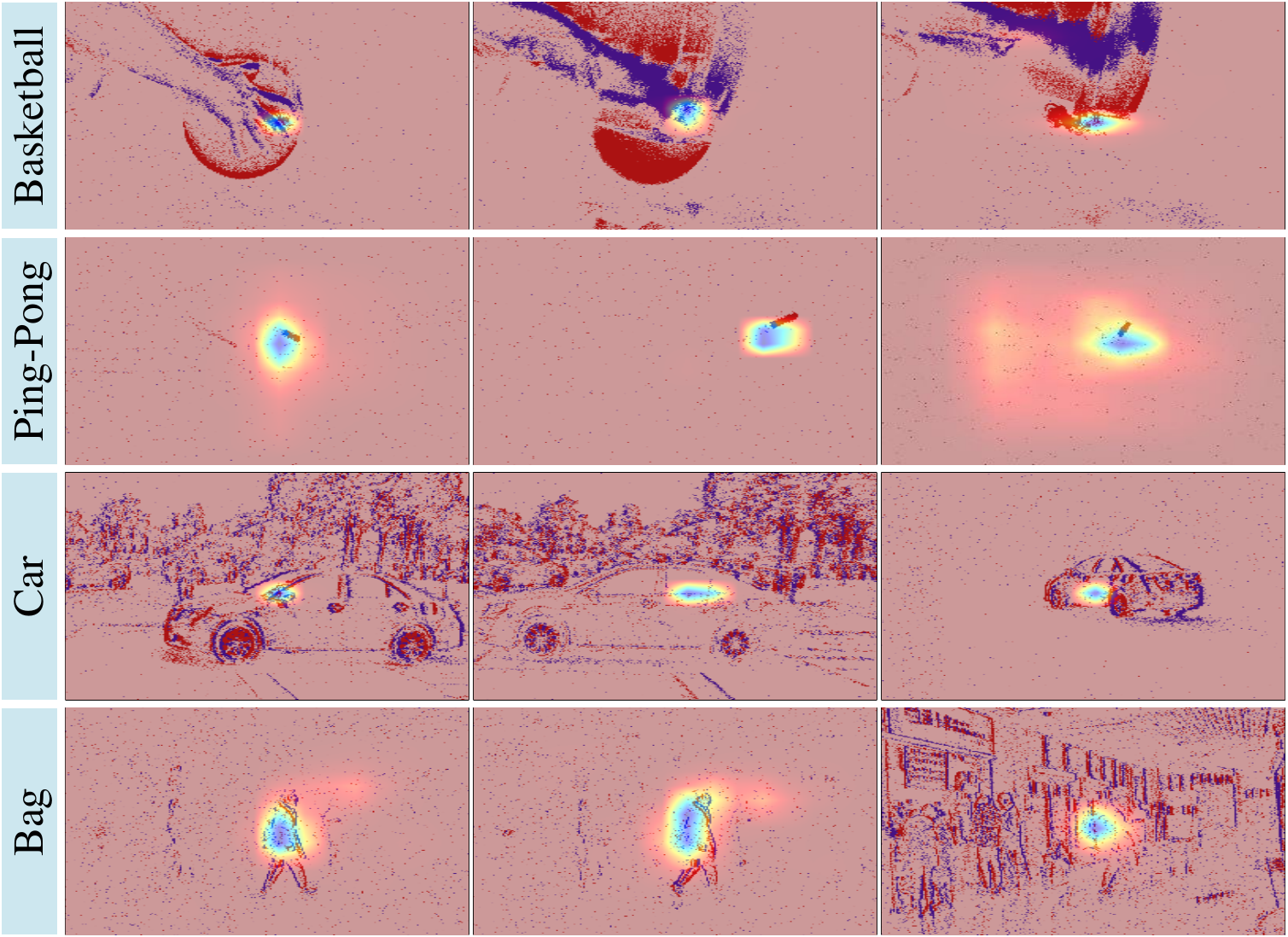}
\caption{Response maps predicted by our method.}  
\label{responseMaps}
\end{figure}

\subsection{Efficiency Analysis} 
In this section, we further conduct an efficiency analysis on the EventVOT dataset. As shown in Table~\ref{CAResults}, introducing SA-MoE leads to the best tracking accuracy, but also increases the model size to 108.5M parameters and reduces the tracking speed to 56 FPS, reflecting the additional computational cost introduced by this module. With the proposed Dynamic Pondering Strategy (DPS), the model maintains comparable accuracy while significantly improving the inference speed to 79 FPS without increasing the parameter count.
Meanwhile, as shown in Table~\ref{EventVOTtable}, we compare the tracking speed of our method with all benchmarked approaches. The results show that our method achieves the best overall tracking accuracy and ranks among the top-performing methods in tracking speed. These efficiency analyses demonstrate that DPS effectively reduces redundant computation by adaptively allocating inference depth, thereby achieving a favorable trade-off between tracking accuracy and efficiency.

\subsection{Visualization}

In addition to the quantitative analysis presented above, we provide qualitative visualizations to offer a more intuitive understanding of the proposed tracking framework.

\noindent $\bullet$ \textbf{Tracking Results.}
As illustrated in Fig.~\ref{trackingResults}, we visualize the tracking results of the proposed PSMTrack and several state-of-the-art trackers, including OSTrack, HDETrack, LMTrack, MCITrack, and ODTrack. In various challenging scenarios, PSMTrack predicts bounding boxes that are more closely aligned with the ground truth than those of competing methods. These visualizations further demonstrate the superior stability and robustness of the proposed method in complex real-world tracking scenarios.

\noindent $\bullet$ \textbf{Attention Heatmaps and Response Maps.}
As illustrated in Fig.~\ref{attention_map}, we visualize the attention heatmaps generated by the proposed method. In these maps, redder regions indicate higher attention weights, suggesting that the model focuses more strongly on the corresponding areas. We can find that the proposed method consistently concentrates on the target object throughout the tracking process, effectively highlighting discriminative target features while suppressing background interference.
Fig.~\ref{responseMaps} further presents the response maps produced by the proposed method, which reflect its localization capability. In these maps, bluer regions correspond to higher response scores. As shown, the proposed method produces concentrated responses on the target, enabling precise and reliable localization.
Together, these visualizations provide qualitative evidence for the robustness and effectiveness of the proposed approach.

\subsection{Limitation Analysis and Feature work} 
Despite achieving a favorable trade-off between tracking accuracy and efficiency, the proposed event-based tracker does not yet incorporate language guidance or exploit the contextual modeling capabilities of multimodal large models. Such semantic cues could provide richer target-level understanding and further improve robustness in complex scenarios. Integrating language descriptions may help disambiguate targets more effectively, while leveraging multimodal models can facilitate visual-textual reasoning and improve localization precision. In future work, we plan to explore how multimodal large models can assist human experts in generating language descriptions for event-based tracking datasets, and how such textual cues can be seamlessly integrated into our framework. This direction is expected to extend our method to more challenging environments with improved adaptability and accuracy.

\section{Conclusion} 
In this paper, we propose PSMTrack, a dynamic pondering sparsity-aware Mixture-of-Experts Transformer for efficient event-based visual tracking. To fully exploit the intrinsic spatial sparsity and temporal density of event data, PSMTrack constructs multi-sparsity event representations from different temporal windows and progressively incorporates dense, medium-density, and sparse search tokens into a three-stage Vision Transformer backbone for hierarchical feature learning. In addition, a sparsity-aware MoE module is introduced to explicitly model event representations with different sparsity levels through expert specialization, thereby enhancing feature extraction and feature discriminability. To further improve efficiency, a dynamic pondering strategy is designed to adaptively adjust the inference depth according to tracking difficulty, reducing redundant computation while maintaining reliable tracking performance. Extensive experiments on the FE240hz, COESOT, and EventVOT datasets demonstrate that PSMTrack achieves competitive tracking accuracy and a favorable trade-off between tracking accuracy and computational efficiency.


\small{ 
\bibliographystyle{IEEEtran}
\bibliography{reference}

\begin{thebibliography}{10}
\providecommand{\url}[1]{#1}
\csname url@samestyle\endcsname
\providecommand{\newblock}{\relax}
\providecommand{\bibinfo}[2]{#2}
\providecommand{\BIBentrySTDinterwordspacing}{\spaceskip=0pt\relax}
\providecommand{\BIBentryALTinterwordstretchfactor}{4}
\providecommand{\BIBentryALTinterwordspacing}{\spaceskip=\fontdimen2\font plus
\BIBentryALTinterwordstretchfactor\fontdimen3\font minus
  \fontdimen4\font\relax}
\providecommand{\BIBforeignlanguage}[2]{{%
\expandafter\ifx\csname l@#1\endcsname\relax
\typeout{** WARNING: IEEEtran.bst: No hyphenation pattern has been}%
\typeout{** loaded for the language `#1'. Using the pattern for}%
\typeout{** the default language instead.}%
\else
\language=\csname l@#1\endcsname
\fi
#2}}
\providecommand{\BIBdecl}{\relax}
\BIBdecl

\bibitem{chen2021transt}
X.~Chen, J.~Yan, Bin~Zhu, D.~Wang, X.~Yang, and H.~Lu, ``Transformer
  tracking,'' in \emph{Proceedings of the IEEE/CVF Conference on Computer
  Vision and Pattern Recognition}, 2021, p. 8126–8135.

\bibitem{wang2021TNL2K}
X.~Wang, X.~Shu, Z.~Zhang, B.~Jiang, Y.~Wang, Y.~Tian, and F.~Wu, ``Towards
  more flexible and accurate object tracking with natural language: Algorithms
  and benchmark,'' in \emph{Proceedings of the IEEE/CVF conference on Computer
  Vision and Pattern Recognition}, 2021, pp. 13\,763--13\,773.

\bibitem{yao2025unctrack}
S.~Yao, Y.~Guo, Y.~Yan, W.~Ren, and X.~Cao, ``Unctrack: Reliable visual object
  tracking with uncertainty-aware prototype memory network,'' \emph{IEEE
  Transactions on Image Processing}, vol.~34, pp. 3533--3546, 2025.

\bibitem{chen2025hyperspectral}
Y.~Chen, Q.~Yuan, H.~Xie, Y.~Tang, Y.~Xiao, J.~He, R.~Guan, X.~Liu, and
  L.~Zhang, ``Hyperspectral video tracking with spectral–spatial fusion and
  memory enhancement,'' \emph{IEEE Transactions on Image Processing}, vol.~34,
  pp. 3547--3562, 2025.

\bibitem{wang2025ssf}
H.~Wang, W.~Li, X.-G. Xia, Q.~Du, and J.~Tian, ``Ssf-net: Spatial-spectral
  fusion network with spectral angle awareness for hyperspectral object
  tracking,'' \emph{IEEE Transactions on Image Processing}, vol.~34, pp.
  3518--3532, 2025.

\bibitem{gallego2020event}
G.~Gallego, T.~Delbr{\"u}ck, G.~Orchard, C.~Bartolozzi, B.~Taba, A.~Censi,
  S.~Leutenegger, A.~J. Davison, J.~Conradt, K.~Daniilidis \emph{et~al.},
  ``Event-based vision: A survey,'' \emph{IEEE Transactions on Pattern Analysis
  and Machine Intelligence}, vol.~44, no.~1, pp. 154--180, 2020.

\bibitem{mambaevt2025}
X.~Wang, C.~Wang, S.~Wang, X.~Wang, Z.~Zhao, L.~Zhu, and B.~Jiang, ``Mambaevt:
  Event stream based visual object tracking using state space model,''
  \emph{IEEE Transactions on Circuits and Systems for Video Technology},
  vol.~36, pp. 278--291, 2026.

\bibitem{zhang2021object}
J.~Zhang, X.~Yang, Y.~Fu, X.~Wei, B.~Yin, and B.~Dong, ``Object tracking by
  jointly exploiting frame and event domain,'' in \emph{Proceedings of the
  IEEE/CVF International Conference on Computer Vision}, 2021, pp.
  13\,043--13\,052.

\bibitem{wang2024event}
X.~Wang, S.~Wang, C.~Tang, L.~Zhu, B.~Jiang, Y.~Tian, and J.~Tang, ``Event
  stream-based visual object tracking: A high-resolution benchmark dataset and
  a novel baseline,'' in \emph{Proceedings of the IEEE/CVF Conference on
  Computer Vision and Pattern Recognition}, 2024, pp. 19\,248--19\,257.

\bibitem{zhu2022learning}
Z.~Zhu, J.~Hou, and X.~Lyu, ``Learning graph-embedded key-event back-tracing
  for object tracking in event clouds,'' \emph{Advances in Neural Information
  Processing Systems}, vol.~35, pp. 7462--7476, 2022.

\bibitem{zhang2026efficient}
J.~Zhang, X.~Yang, H.~Tang, Y.~Wang, B.~Yin, H.~Wang, and X.~Fu, ``Efficient
  vision transformer with token sparsification for event-based object
  tracking,'' \emph{International Journal of Computer Vision}, vol. 134, no.~2,
  p.~75, 2026.

\bibitem{dosovitskiy2020image}
A.~Dosovitskiy, L.~Beyer, A.~Kolesnikov, D.~Weissenborn, X.~Zhai,
  T.~Unterthiner, M.~Dehghani, M.~Minderer, G.~Heigold, S.~Gelly \emph{et~al.},
  ``An image is worth 16x16 words: Transformers for image recognition at
  scale,'' \emph{arXiv preprint arXiv:2010.11929}, 2020.

\bibitem{marvasti2021trackSurvey}
S.~M. Marvasti-Zadeh, L.~Cheng, H.~Ghanei-Yakhdan, and S.~Kasaei, ``Deep
  learning for visual tracking: A comprehensive survey,'' \emph{IEEE
  Transactions on Intelligent Transportation Systems}, vol.~23, no.~5, pp.
  3943--3968, 2021.

\bibitem{8368143}
J.~Huang, S.~Wang, M.~Guo, and S.~Chen, ``Event-guided structured output
  tracking of fast-moving objects using a celex sensor,'' \emph{IEEE
  Transactions on Circuits and Systems for Video Technology}, vol.~28, no.~9,
  pp. 2413--2417, 2018.

\bibitem{chen2019asynchronous}
H.~Chen, Q.~Wu, Y.~Liang, X.~Gao, and H.~Wang, ``Asynchronous
  tracking-by-detection on adaptive time surfaces for event-based object
  tracking,'' in \emph{Proceedings of the 27th ACM International Conference on
  Multimedia}, 2019, pp. 473--481.

\bibitem{gehrig2020eklt}
D.~Gehrig, H.~Rebecq, G.~Gallego, and D.~Scaramuzza, ``Eklt: Asynchronous
  photometric feature tracking using events and frames,'' \emph{International
  Journal of Computer Vision}, vol. 128, no.~3, pp. 601--618, 2020.

\bibitem{zhang2022spiking}
J.~Zhang, B.~Dong, H.~Zhang, J.~Ding, F.~Heide, B.~Yin, and X.~Yang, ``Spiking
  transformers for event-based single object tracking,'' in \emph{Proceedings
  of the IEEE/CVF conference on Computer Vision and Pattern Recognition}, 2022,
  pp. 8801--8810.

\bibitem{wang2025towards}
S.~Wang, X.~Wang, L.~Jin, B.~Jiang, L.~Zhu, L.~Chen, Y.~Tian, and B.~Luo,
  ``Towards low-latency event stream-based visual object tracking: A slow-fast
  approach,'' \emph{arXiv preprint arXiv:2505.12903}, 2025.

\bibitem{yan2021stark}
B.~Yan, H.~Peng, J.~Fu, D.~Wang, and H.~Lu, ``Learning spatio-temporal
  transformer for visual tracking,'' in \emph{Proceedings of the IEEE/CVF
  International Conference on Computer Vision}, 2021, p. 10448–10457.

\bibitem{ye2022ostrack}
B.~Ye, H.~Chang, B.~Ma, S.~Shan, and X.~Chen, ``Joint feature learning and
  relation modeling for tracking: A one-stream framework,'' in \emph{European
  conference on computer vision}.\hskip 1em plus 0.5em minus 0.4em\relax
  Springer, 2022, pp. 341--357.

\bibitem{zhu2025two}
J.~Zhu, H.~Tang, X.~Chen, X.~Wang, D.~Wang, and H.~Lu, ``Two-stream beats
  one-stream: Asymmetric siamese network for efficient visual tracking,'' in
  \emph{Proceedings of the AAAI Conference on Artificial Intelligence},
  vol.~39, no.~10, 2025, pp. 10\,959--10\,967.

\bibitem{hong2025general}
L.~Hong, J.~Li, X.~Zhou, S.~Yan, P.~Guo, K.~Jiang, Z.~Chen, S.~Gao, R.~Li,
  X.~Sheng \emph{et~al.}, ``General compression framework for efficient
  transformer object tracking,'' in \emph{Proceedings of the IEEE/CVF
  International Conference on Computer Vision}, 2025, pp. 13\,427--13\,437.

\bibitem{xu2025less}
C.~Xu, B.~Zhong, Q.~Liang, Y.~Zheng, G.~Li, and S.~Song, ``Less is more: Token
  context-aware learning for object tracking,'' in \emph{Proceedings of the
  AAAI Conference on Artificial Intelligence}, vol.~39, no.~8, 2025, pp.
  8824--8832.

\bibitem{xue2025similarity}
C.~Xue, B.~Zhong, Q.~Liang, Y.~Zheng, N.~Li, Y.~Xue, and S.~Song,
  ``Similarity-guided layer-adaptive vision transformer for uav tracking,'' in
  \emph{Proceedings of the Computer Vision and Pattern Recognition Conference},
  2025, pp. 6730--6740.

\bibitem{yan2021lighttrack}
B.~Yan, H.~Peng, K.~Wu, D.~Wang, J.~Fu, and H.~Lu, ``Lighttrack: Finding
  lightweight neural networks for object tracking via one-shot architecture
  search,'' in \emph{Proceedings of the IEEE/CVF conference on Computer Vision
  and Pattern Recognition}, 2021, pp. 15\,180--15\,189.

\bibitem{cui2023mixformerv2}
Y.~Cui, T.~Song, G.~Wu, and L.~Wang, ``Mixformerv2: Efficient fully transformer
  tracking,'' \emph{Advances in Neural Information Processing Systems},
  vol.~36, pp. 58\,736--58\,751, 2023.

\bibitem{zhu2025exploring}
J.~Zhu, X.~Chen, H.~Diao, S.~Li, J.-Y. He, C.~Li, B.~Luo, D.~Wang, and H.~Lu,
  ``Exploring dynamic transformer for efficient object tracking,'' \emph{IEEE
  Transactions on Neural Networks and Learning Systems}, vol.~36, no.~8, pp.
  15\,502--15\,514, 2025.

\bibitem{yin2022vit}
H.~Yin, A.~Vahdat, J.~M. Alvarez, A.~Mallya, J.~Kautz, and P.~Molchanov,
  ``A-vit: Adaptive tokens for efficient vision transformer,'' in
  \emph{Proceedings of the IEEE/CVF conference on Computer Vision and Pattern
  Recognition}, 2022, pp. 10\,809--10\,818.

\bibitem{shazeer2017outrageously}
N.~Shazeer, A.~Mirhoseini, K.~Maziarz, A.~Davis, Q.~V. Le, G.~E. Hinton, and
  J.~Dean, ``Outrageously large neural networks: The sparsely-gated
  mixture-of-experts layer,'' in \emph{International Conference on Learning
  Representations}, 2017.

\bibitem{fedus2022switch}
W.~Fedus, B.~Zoph, and N.~Shazeer, ``Switch transformers: Scaling to trillion
  parameter models with simple and efficient sparsity,'' \emph{Journal of
  Machine Learning Research}, vol.~23, no. 120, pp. 1--39, 2022.

\bibitem{riquelme2021scaling}
C.~Riquelme, J.~Puigcerver, B.~Mustafa, M.~Neumann, R.~Jenatton,
  A.~Susano~Pinto, D.~Keysers, and N.~Houlsby, ``Scaling vision with sparse
  mixture of experts,'' \emph{Advances in Neural Information Processing
  Systems}, vol.~34, pp. 8583--8595, 2021.

\bibitem{hwang2023tutel}
C.~Hwang, W.~Cui, Y.~Xiong, Z.~Yang, Z.~Liu, H.~Hu, Z.~Wang, R.~Salas, J.~Jose,
  P.~Ram \emph{et~al.}, ``Tutel: Adaptive mixture-of-experts at scale,''
  \emph{Proceedings of Machine Learning and Systems}, vol.~5, pp. 269--287,
  2023.

\bibitem{lewis2021base}
M.~Lewis, S.~Bhosale, T.~Dettmers, N.~Goyal, and L.~Zettlemoyer, ``Base layers:
  Simplifying training of large, sparse models,'' in \emph{International
  Conference on Machine Learning}, 2021, pp. 6265--6274.

\bibitem{zoph2022stmoe}
B.~Zoph, I.~Bello, S.~Kumar, N.~Du, Y.~Huang, J.~Dean, N.~Shazeer, and
  W.~Fedus, ``St-moe: Designing stable and transferable sparse expert models,''
  in \emph{International Conference on Learning Representations}, 2022.

\bibitem{roller2021hash}
S.~Roller, S.~Sukhbaatar, J.~Weston \emph{et~al.}, ``Hash layers for large
  sparse models,'' \emph{Advances in Neural Information Processing Systems},
  vol.~34, pp. 17\,555--17\,566, 2021.

\bibitem{lu2025dynamic}
Y.~Lu, M.~Weng, Z.~Xiao, R.~Jiang, W.~Su, G.~Zheng, P.~Lu, and X.~Li,
  ``Dynamic-dino: Fine-grained mixture of experts tuning for real-time
  open-vocabulary object detection,'' in \emph{Proceedings of the IEEE/CVF
  International Conference on Computer Vision}, 2025, pp. 20\,847--20\,856.

\bibitem{zhao2025equipping}
S.~Zhao, J.~Liu, X.~Wen, H.~Tan, and X.~Qi, ``Equipping vision foundation model
  with mixture of experts for out-of-distribution detection,'' in
  \emph{Proceedings of the IEEE/CVF International Conference on Computer
  Vision}, 2025, pp. 1751--1761.

\bibitem{zhu2025separation}
Y.~Zhu, H.~Chen, Y.~Deng, and W.~You, ``Separation for better integration:
  Disentangling edge and motion in event-based deblurring,'' in
  \emph{Proceedings of the IEEE/CVF International Conference on Computer
  Vision}, 2025, pp. 14\,732--14\,742.

\bibitem{graves2016adaptive}
A.~Graves, ``Adaptive computation time for recurrent neural networks,''
  \emph{arXiv preprint arXiv:1603.08983}, 2016.

\bibitem{xu2020siamfc++}
Y.~Xu, Z.~Wang, Z.~Li, Y.~Yuan, and G.~Yu, ``Siamfc++: Towards robust and
  accurate visual tracking with target estimation guidelines,'' in
  \emph{Proceedings of the AAAI Conference on Artificial Intelligence},
  vol.~34, no.~07, 2020, pp. 12\,549--12\,556.

\bibitem{goutam2019Dimp}
G.~Bhat, M.~Danelljan, L.~V. Gool, and R.~Timofte, ``Learning discriminative
  model prediction for tracking,'' in \emph{Proceedings of the IEEE/CVF
  International Conference on Computer Vision}, 2019, p. 6182–6191.

\bibitem{danelljan2019atom}
M.~Danelljan, G.~Bhat, F.~S. Khan, and M.~Felsberg, ``Atom: Accurate tracking
  by overlap maximization,'' in \emph{Proceedings of the IEEE/CVF conference on
  Computer Vision and Pattern Recognition}, 2019, pp. 4660--4669.

\bibitem{cui2022mixformer}
Y.~Cui, C.~Jiang, L.~Wang, and W.~Gangshan, ``Mixformer: End-to-end tracking
  with iterative mixed attention,'' in \emph{Proceedings of the IEEE/CVF
  Conference on Computer Vision and Pattern Recognition}, 2022, p.
  13608–13618.

\bibitem{xie2024autoregressive}
J.~Xie, B.~Zhong, Z.~Mo, S.~Zhang, L.~Shi, S.~Song, and R.~Ji, ``Autoregressive
  queries for adaptive tracking with spatio-temporal transformers,'' in
  \emph{Proceedings of the IEEE/CVF Conference on Computer Vision and Pattern
  Recognition}, 2024, pp. 19\,300--19\,309.

\bibitem{zhang2021fe108}
J.~Zhang, X.~Yang, Y.~Fu, X.~Wei, B.~Yin, and B.~Dong, ``Object tracking by
  jointly exploiting frame and event domain,'' in \emph{Proceedings of the
  IEEE/CVF International Conference on Computer Vision}, 2021, pp.
  13\,043--13\,052.

\bibitem{tang2025revisiting}
C.~Tang, X.~Wang, J.~Huang, B.~Jiang, L.~Zhu, S.~Chen, J.~Zhang, Y.~Wang, and
  Y.~Tian, ``Revisiting color-event based tracking: A unified network, dataset,
  and metric,'' \emph{Pattern Recognition}, p. 112718, 2025.

\bibitem{loshchilov2018adamw}
I.~Loshchilov and F.~Hutter, ``Decoupled weight decay regularization,'' in
  \emph{International Conference on Learning Representations}, 2018.

\bibitem{paszke2019pytorch}
A.~Paszke, S.~Gross, F.~Massa, A.~Lerer, J.~Bradbury, G.~Chanan, T.~Killeen,
  Z.~Lin, N.~Gimelshein, L.~Antiga \emph{et~al.}, ``Pytorch: An imperative
  style, high-performance deep learning library,'' \emph{Advances in Neural
  Information Processing Systems}, vol.~32, 2019.

\bibitem{wang2021TrDiMP}
N.~Wang, W.~Zhou, J.~Wang, and H.~Li, ``Transformer meets tracker: Exploiting
  temporal context for robust visual tracking,'' in \emph{Proceedings of the
  IEEE/CVF Conference on Computer Vision and Pattern Recognition}, 2021, p.
  1571–1580.

\bibitem{mayer2022Tomp}
C.~Mayer, M.~Danelljan, G.~Bhat, M.~Paul, D.~P. Paudel, F.~Yu, and L.~V. Gool,
  ``Transforming model prediction for tracking,'' in \emph{Proceedings of the
  IEEE/CVF Conference on Computer Vision and Pattern Recognition}, 2022, p.
  8731–8740.

\bibitem{gao2022AIa}
S.~Gao, C.~Zhou, C.~Ma, X.~Wang, and J.~Yuan, ``Aiatrack: Attention in
  attention for transformer visual tracking,'' in \emph{European Conference on
  Computer Vision}, 2022, p. 146–164.

\bibitem{martin2020PrDimp}
M.~Danelljan, L.~V. Gool, and R.~Timofte, ``Probabilistic regression for visual
  tracking,'' in \emph{Proceedings of the IEEE/CVF Conference on Computer
  Vision and Pattern Recognition}, 2019, p. 7183–7192.

\bibitem{bhat2022SKys}
G.~Bhat, M.~Danelljan, L.~Van~Gool, and R.~Timofte, ``Know your surroundings:
  Exploiting scene information for object tracking,'' in \emph{European
  Conference on Computer Vision}, 2020, p. 205–221.

\bibitem{chen2022simtrack}
B.~Chen, P.~Li, L.~Bai, L.~Qiao, Q.~Shen, B.~Li, W.~Gan, W.~Wu, and W.~Ouyang,
  ``Backbone is all your need: A simplified architecture for visual object
  tracking,'' in \emph{European Conference on Computer Vision}, 2021, p.
  375–392.

\bibitem{kang2025exploring}
B.~Kang, X.~Chen, S.~Lai, Y.~Liu, Y.~Liu, and D.~Wang, ``Exploring enhanced
  contextual information for video-level object tracking,'' in
  \emph{Proceedings of the AAAI Conference on Artificial Intelligence},
  vol.~39, no.~4, 2025, pp. 4194--4202.

\bibitem{wu2026utptrack}
H.~Wu, X.~Wang, J.~Zhang, J.~Tong, X.~Chen, J.~Lin, Y.~Ma, and X.~Shen,
  ``Utptrack: Towards simple and unified token pruning for visual tracking,''
  \emph{arXiv preprint arXiv:2602.23734}, 2026.

\bibitem{zhang2026spiketrack}
Q.~Zhang, J.~Cheng, Q.~Mao, C.~Liu, Y.~Fang, Y.~Li, M.~Ge, and S.~Gao,
  ``Spiketrack: A spike-driven framework for efficient visual tracking,''
  \emph{arXiv preprint arXiv:2602.23963}, 2026.

\bibitem{martin2019Atom}
M.~Danelljan, G.~Bhat, F.~Shahbaz~Khan, and M.~Felsberg, ``Atom: Accurate
  tracking by overlap maximization,'' in \emph{Proceedings of the IEEE/CVF
  Conference on Computer Vision and Pattern Recognition}, 2019, p. 4660–4669.

\bibitem{cai2023robust}
Y.~Cai, J.~Liu, J.~Tang, and G.~Wu, ``Robust object modeling for visual
  tracking,'' in \emph{Proceedings of the IEEE/CVF International Conference on
  Computer Vision}, 2023, pp. 9589--9600.

\bibitem{wei2023autoregressive}
X.~Wei, Y.~Bai, Y.~Zheng, D.~Shi, and Y.~Gong, ``Autoregressive visual
  tracking,'' in \emph{Proceedings of the IEEE/CVF Conference on Computer
  Vision and Pattern Recognition}, 2023, pp. 9697--9706.

\bibitem{zheng2024odtrack}
Y.~Zheng, B.~Zhong, Q.~Liang, Z.~Mo, S.~Zhang, and X.~Li, ``Odtrack: Online
  dense temporal token learning for visual tracking,'' in \emph{Proceedings of
  the AAAI Conference on Artificial Intelligence}, vol.~38, no.~7, 2024, pp.
  7588--7596.

\bibitem{shi2024explicit}
L.~Shi, B.~Zhong, Q.~Liang, N.~Li, S.~Zhang, and X.~Li, ``Explicit visual
  prompts for visual object tracking,'' in \emph{Proceedings of the AAAI
  Conference on Artificial Intelligence}, vol.~38, no.~5, 2024, pp. 4838--4846.

\bibitem{bai2024artrackv2}
Y.~Bai, Z.~Zhao, Y.~Gong, and X.~Wei, ``Artrackv2: Prompting autoregressive
  tracker where to look and how to describe,'' in \emph{Proceedings of the
  IEEE/CVF Conference on Computer Vision and Pattern Recognition}, 2024, pp.
  19\,048--19\,057.

\end{thebibliography}
}

\end{document}